# ECA: High Dimensional Elliptical Component Analysis in non-Gaussian Distributions

Fang Han* and Han Liu‡

**Abstract**

We present a robust alternative to principal component analysis (PCA) — called elliptical component analysis (ECA) — for analyzing high dimensional, elliptically distributed data. ECA estimates the eigenspace of the covariance matrix of the elliptical data. To cope with heavy-tailed elliptical distributions, a multivariate rank statistic is exploited. At the model-level, we consider two settings: either that the leading eigenvectors of the covariance matrix are non-sparse or that they are sparse. Methodologically, we propose ECA procedures for both non-sparse and sparse settings. Theoretically, we provide both non-asymptotic and asymptotic analyses quantifying the theoretical performances of ECA. In the non-sparse setting, we show that ECA's performance is highly related to the effective rank of the covariance matrix. In the sparse setting, the results are twofold: (i) We show that the sparse ECA estimator based on a combinatoric program attains the optimal rate of convergence; (ii) Based on some recent developments in estimating sparse leading eigenvectors, we show that a computationally efficient sparse ECA estimator attains the optimal rate of convergence under a suboptimal scaling.

## 1 Introduction

Principal component analysis (PCA) plays important roles in many different areas. For example, it is one of the most useful techniques for data visualization in studying brain imaging data (Lindquist, 2008). This paper considers a problem closely related to PCA, namely estimating the leading eigenvectors of the covariance matrix. Let $\boldsymbol{X}_1, \ldots, \boldsymbol{X}_n$ be $n$ data points of a random vector $\boldsymbol{X} \in \mathbb{R}^d$. Denote $\boldsymbol{\Sigma}$ to be the covariance matrix of $\boldsymbol{X}$, and $\boldsymbol{u}_1, \ldots, \boldsymbol{u}_m$ to be its top $m$ leading eigenvectors. We want to find $\widehat{\boldsymbol{u}}_1, \ldots, \widehat{\boldsymbol{u}}_m$ that can estimate $\boldsymbol{u}_1, \ldots, \boldsymbol{u}_m$ accurately.

*Department of Statistics, University of Washington, Seattle, WA 98115, USA; e-mail: fanghan@uw.edu

†Department of Operations Research and Financial Engineering, Princeton University, Princeton, NJ 08544, USA; e-mail: hanliu@princeton.edu

‡Fang Han's research was supported by NIH-R01-EB012547 and a UW faculty start-up grant. Han Liu's research was supported by the NSF CAREER Award DMS-1454377, NSF IIS-1546482, NSF IIS-1408910, NSF IIS-1332109, NIH R01-MH102339, NIH R01-GM083084, and NIH R01-HG06841.

This paper is focused on the high dimensional setting where the dimension $d$ could be comparable to, or even larger than, the sample size $n$, and the data could be heavy-tailed (especially, non-Gaussian). A motivating example for considering such high dimensional non-Gaussian heavy-tailed data is our study on functional magnetic resonance imaging (fMRI). In particular, in Section 6 we examine an fMRI data, with 116 regions of interest (ROIs), from the Autism Brian Imaging Data Exchange (ABIDE) project containing 544 normal subjects. There, the dimension $d = 116$ is comparable to the sample size $n = 544$. In addition, Table 3 shows that the data we consider cannot pass any normality test, and Figure 5 further indicates that the data are heavy-tailed.

In high dimensions, the performance of PCA, using the leading eigenvectors of the Pearson's sample covariance matrix, has been studied for subgaussian data. In particular, for any matrix $\mathbf{M} \in \mathbb{R}^{d\times d}$, letting $\mathrm{Tr}(\mathbf{M})$ and $\sigma_i(\mathbf{M})$ be the trace and $i$-th largest singular value of $\mathbf{M}$, Lounici (2014) showed that PCA could be consistent when $r^*(\boldsymbol{\Sigma}) := \mathrm{Tr}(\boldsymbol{\Sigma})/\sigma_1(\boldsymbol{\Sigma})$ satisfies $r^*(\boldsymbol{\Sigma}) \log d/n \to 0$. $r^*(\boldsymbol{\Sigma})$ is referred to as the effective rank of $\boldsymbol{\Sigma}$ in the literature (Vershynin, 2010; Lounici, 2014).

When $r^*(\boldsymbol{\Sigma}) \log d/n \not\to 0$, PCA might not be consistent. The inconsistency phenomenon of PCA in high dimensions has been pointed out by Johnstone and Lu (2009). In particular, they showed that the angle between the PCA estimator and $\boldsymbol{u}_1$ may not converge to 0 if $d/n \to c$ for some constant $c > 0$. To avoid this curse of dimensionality, certain types of sparsity assumptions are needed. For example, in estimating the leading eigenvector $\boldsymbol{u}_1 := (u_{11}, \ldots, u_{1d})^T$, we may assume that $\boldsymbol{u}_1$ is sparse, i.e., $s := \mathrm{card}(\{j : u_{1j} \neq 0\}) \ll n$. We call the setting that $\boldsymbol{u}_1$ is sparse the "sparse setting" and the setting that $\boldsymbol{u}_1$ is not necessarily sparse the "non-sparse setting".

In the sparse setting, different variants of sparse PCA methods have been proposed. For example, d'Aspremont et al. (2007) proposed formulating a convex semidefinite program for calculating the sparse leading eigenvectors. Jolliffe et al. (2003) and Zou et al. (2006) connected PCA to regression and proposed using lasso-type estimators for parameter estimation. Shen and Huang (2008) and Witten et al. (2009) connected PCA to singular vector decomposition (SVD) and proposed iterative algorithms for estimating the left and right singular vectors. Journée et al. (2010) and Zhang and El Ghaoui (2011) proposed greedily searching the principal submatrices of the covariance matrix. Ma (2013) and Yuan and Zhang (2013) proposed using modified versions of the power method to estimate eigenvectors and principal subspaces.

Theoretical properties of these methods have been analyzed under both Gaussian and subgaussian assumptions. On one hand, in terms of computationally efficient methods, under the spike covariance Gaussian model, Amini and Wainwright (2009) showed the consistency in parameter estimation and model selection for sparse PCA computed via the semidefinite program proposed in d'Aspremont et al. (2007). Ma (2013) justified the use of a modified iterative thresholding method in estimating principal subspaces. By exploiting a convex program using the Fantope projection (Overton and Womersley, 1992; Dattorro, 2005), Vu et al. (2013) showed that there exist computationally efficient estimators that attain (under various settings) $O_P(s\sqrt{\log d/n})$ rate of convergence for possibly non-spike model. In Section 5.1, we will discuss the Fantope projection in more detail.

On the other hand, there exists another line of research focusing on studying sparse PCA conducted via combinatoric programs. For example, Vu and Lei (2012), Lounici (2013), and Vu and Lei (2013) studied leading eigenvector and principal subspace estimation problems via exhaustively searching over all submatrices. They showed that the optimal $O_P(\sqrt{s\log(ed/s)/n})$ (up to some other parameters of $\mathbf{\Sigma}$) rate of convergence can be attained using this computationally expensive approach. Such a global search was also studied in Cai et al. (2015), where they established the upper and lower bounds in both covariance matrix and principal subspace estimations. Barriers between the aforementioned statistically efficient method and computationally efficient methods in sparse PCA was pointed out by Berthet and Rigollet (2013) using the principal component detection problem. Such barriers were also studied in Ma and Wu (2015).

One limitation for the PCA and sparse PCA theories is that they rely heavily on the Gaussian or subgaussian assumption. If the Gaussian assumption is correct, accurate estimation can be expected, otherwise, the obtained result may be misleading. To relax the Gaussian assumption, Han and Liu (2014) generalized the Gaussian to the semiparametric transelliptical family (called the "meta-elliptical" in their paper) for modeling the data. The transelliptical family assumes that, after unspecified increasing marginal transformations, the data are elliptically distributed. By resorting to the marginal Kendall's tau statistic, Han and Liu (2014) proposed a semiparametric alternative to scale-invariant PCA, named transelliptical component analysis (TCA), for estimating the leading eigenvector of the latent generalized correlation matrix $\mathbf{\Sigma}^0$. In follow-up works, Wegkamp and Zhao (2016) and Han and Liu (2016) showed that, under various settings, (i) In the non-sparse case, TCA attains the $O_P(\sqrt{r^*(\mathbf{\Sigma}^0)\log d/n})$ rate of convergence in parameter estimation, which is the same rate of convergence for PCA under the subgaussian assumption (Lounici, 2014; Bunea and Xiao, 2015); (ii) In the sparse case, sparse TCA, formulated as a combinatoric program, can attain the optimal $O_P(\sqrt{s\log(ed/s)/n})$ rate of convergence under the "sign subgaussian" condition. More recently, Vu et al. (2013) showed that, sparse TCA, via the Fantope projection, can attain the $O_P(s\sqrt{\log d/n})$ rate of convergence.

Despite all these efforts, there are two remaining problems for the aforementioned works exploiting the marginal Kendall's tau statistic. First, using marginal ranks, they can only estimate the leading eigenvectors of the correlation matrix instead of the covariance matrix. Secondly, the sign subgaussian condition is not easy to verify.

In this paper we show that, under the elliptical model and various settings (see Corollaries 3.1, 4.1, and Theorems 5.4 and 3.5 for details), the $O_P(\sqrt{s\log(ed/s)/n})$ rate of convergence for estimating the leading eigenvector of $\mathbf{\Sigma}$ can be attained without the need of sign subgaussian condition. In particular, we present an alternative procedure, called elliptical component analysis (ECA), to directly estimate the eigenvectors of $\mathbf{\Sigma}$ and treat the corresponding eigenvalues as nuisance parameters. ECA exploits the multivariate Kendall's tau for estimating the eigenspace of $\mathbf{\Sigma}$. When the target parameter is sparse, the corresponding ECA procedure is called sparse ECA.

We show that (sparse) ECA, under various settings, has the following properties. (i) In the non-sparse setting, ECA attains the efficient $O_P(\sqrt{r^*(\mathbf{\Sigma})\log d/n})$ rate of convergence. (ii) In the sparse setting, sparse ECA, via a combinatoric program, attains the minimax optimal $O_P(\sqrt{s\log(ed/s)/n})$ rate of convergence. (iii) In the sparse setting, sparse ECA, via a computationally efficient program which combines the Fantope projection (Vu et al., 2013) and truncated power algorithm (Yuan and Zhang, 2013), attains the optimal $O_P(\sqrt{s\log(ed/s)/n})$ rate of convergence under a suboptimal scaling ($s^2\log d/n \to 0$). Of note, for presentation clearness, the rates presented here omit a variety of parameters regarding $\mathbf{\Sigma}$ and $\mathbf{K}$. The readers should refer to Corollaries 3.1, 4.1, and Theorem 5.4 for accurate descriptions.

Table 1: The illustration of the results in (sparse) PCA, (sparse) TCA, and (sparse) ECA for the leading eigenvector estimation. Similar results also hold for principal subspace estimation. Here $\mathbf{\Sigma}$ is the covariance matrix, $\mathbf{\Sigma}^0$ is the latent generalized correlation matrix, $r^*(\mathbf{M}) := \text{Tr}(\mathbf{M})/\sigma_1(\mathbf{M})$ represents the effective rank of $\mathbf{M}$, "r.c." stands for "rate of convergence", "n-s setting" stands for the "non-sparse setting", "sparse setting 1" stands for the "sparse setting" where the estimation procedure is conducted via a combinatoric program, "sparse setting 2" stands for the "sparse setting" where the estimation procedure is conducted via combining the Fantope projection (Vu et al., 2013) and the truncated power method (Yuan and Zhang, 2013). For presentation clearness, the rates presented here omit a variety of parameters regarding $\mathbf{\Sigma}$ and $\mathbf{K}$. The readers should refer to Corollaries 3.1, 4.1, and Theorem 5.4 for accurate descriptions.

| | (sparse) PCA | (sparse) TCA | (sparse) ECA |
|---|---|---|---|
| working model: | subgaussian family | transelliptical family | elliptical family |
| parameter of interest: | eigenvectors of $\mathbf{\Sigma}$ | eigenvectors of $\mathbf{\Sigma}^0$ | eigenvectors of $\mathbf{\Sigma}$ |
| input statistics: | Pearson's covariance matrix | Kendall's tau | multivariate Kendall's tau |
| n-s setting (r.c.): | $\sqrt{r^*(\mathbf{\Sigma})\log d/n}$ | $\sqrt{r^*(\mathbf{\Sigma}^0)\log d/n}$ | $\sqrt{r^*(\mathbf{\Sigma})\log d/n}$ |
| sparse setting 1 (r.c): | $\sqrt{s\log(ed/s)/n}$ | $s\sqrt{\log d/n}$ (general),<br>$\sqrt{s\log(ed/s)/n}$ (sign subgaussian) | $\sqrt{s\log(ed/s)/n}$ |
| sparse setting 2 (r,c): | $\sqrt{s\log(ed/s)/n}$<br>given $s^2\log d/n \to 0$ | $s\sqrt{\log d/n}$ (general),<br>$\sqrt{s\log(ed/s)/n}$ (sign subgaussian)<br>given $s^2\log d/n \to 0$ | $\sqrt{s\log(ed/s)/n}$<br>given $s^2\log d/n \to 0$ |

We compare (sparse) PCA, (sparse) TCA, and (sparse) ECA in Table 1.

## 1.1 Related Works

The multivariate Kendall's tau statistic is first introduced in Choi and Marden (1998) for testing independence and is further used in estimating low-dimensional covariance matrices (Visuri et al., 2000; Oja, 2010) and principal components (Marden, 1999; Croux et al., 2002; Jackson and Chen, 2004). In particular, Marden (1999) showed that the population multivariate Kendall's tau, $\mathbf{K}$, shares the same eigenspace as the covariance matrix $\mathbf{\Sigma}$. Croux et al. (2002) illustrated the asymptotical efficiency of ECA compared to PCA for the Gaussian data when $d = 2$ and 3. Taskinen et al. (2012) characterized the robustness and efficiency properties of ECA in low dimensions

Some related methods using multivariate rank-based statistics are discussed in Tyler (1982), Tyler (1987), Taskinen et al. (2003), Oja and Randles (2004), Oja and Paindaveine (2005), Oja et al. (2006), and Sirkiä et al. (2007). Theoretical analysis in low dimensions is provided in Hallin and Paindaveine (2002b,a, 2004, 2005, 2006), Hallin et al. (2006, 2010, 2014), and some new extensions to high dimensional settings are provided in Croux et al. (2013) and Feng (2015).

Our paper has significantly new contributions to high dimensional robust statistics literature. Theoretically, we study the use of the multivariate Kendall's tau in high dimensions, provide new properties of the multivariate rank statistic, and characterize the performance of ECA in both non-sparse and sparse settings. Computationally, we provide an efficient algorithm for conducting sparse ECA and highlight the "optimal rate, suboptimal scaling" phenomenon in understanding the behavior of the proposed algorithm.

### 1.2 Notation

Let $\mathbf{M} = [\mathbf{M}_{jk}] \in \mathbb{R}^{d\times d}$ be a symmetric matrix and $\boldsymbol{v} = (v_1, ..., v_d)^T \in \mathbb{R}^d$ be a vector. We denote $\boldsymbol{v}_I$ to be the subvector of $\boldsymbol{v}$ whose entries are indexed by a set $I$, and $\mathbf{M}_{I,J}$ to be the submatrix of $\mathbf{M}$ whose rows are indexed by $I$ and columns are indexed by $J$. We denote $\mathrm{supp}(\boldsymbol{v}) := \{j : v_j \neq 0\}$. For $0 < q < \infty$, we define the $\ell_q$ and $\ell_\infty$ vector norms as $\|\boldsymbol{v}\|_q := (\sum_{i=1}^d |v_i|^q)^{1/q}$ and $\|\boldsymbol{v}\|_\infty := \max_{1\le i\le d} |v_i|$. We denote $\|\boldsymbol{v}\|_0 := \mathrm{card}(\mathrm{supp}(\boldsymbol{v}))$. We define the matrix entry-wise maximum value and Frobenius norms as $\|\mathbf{M}\|_{\max} := \max\{|\mathbf{M}_{ij}|\}$ and $\|\mathbf{M}\|_{\mathsf{F}} = (\sum \mathbf{M}_{jk}^2)^{1/2}$. Let $\lambda_j(\mathbf{M})$ be the $j$-th largest eigenvalue of $\mathbf{M}$. If there are ties, $\lambda_j(\mathbf{M})$ is any one of the eigenvalues such that any eigenvalue larger than it has rank smaller than $j$, and any eigenvalue smaller than it has rank larger than $j$. Let $\boldsymbol{u}_j(\mathbf{M})$ be any unit vector $\boldsymbol{v}$ such that $\boldsymbol{v}^T\mathbf{M}\boldsymbol{v} = \lambda_j(\mathbf{M})$. Without loss of generality, we assume that the first nonzero entry of $\boldsymbol{u}_j(\mathbf{M})$ is positive. We denote $\|\mathbf{M}\|_2$ to be the spectral norm of $\mathbf{M}$ and $\mathbb{S}^{d-1} := \{\boldsymbol{v} \in \mathbb{R}^d : \|\boldsymbol{v}\|_2 = 1\}$ to be the $d$-dimensional unit sphere. We define the restricted spectral norm $\|\mathbf{M}\|_{2,s} := \sup_{\boldsymbol{v}\in\mathbb{S}^{d-1}, \|\boldsymbol{v}\|_0\le s} |\boldsymbol{v}^T\mathbf{M}\boldsymbol{v}|$, so for $s = d$, we have $\|\mathbf{M}\|_{2,s} = \|\mathbf{M}\|_2$. We denote $f(\mathbf{M})$ to be the matrix with entries $[f(\mathbf{M})]_{jk} = f(\mathbf{M}_{jk})$. We denote $\mathrm{diag}(\mathbf{M})$ to be the diagonal matrix with the same diagonal entries as $\mathbf{M}$. Let $\mathbf{I}_d$ represent the $d$ by $d$ identity matrix. For any two numbers $a, b \in \mathbb{R}$, we denote $a \wedge b := \min\{a, b\}$ and $a \vee b := \max\{a, b\}$. For any two sequences of positive numbers $\{a_n\}$ and $\{b_n\}$, we write $a_n \asymp b_n$ if $a_n = O(b_n)$ and $b_n = O(a_n)$. We write $b_n = \Omega(a_n)$ if $a_n = O(b_n)$, and $b_n = \Omega^o(a_n)$ if $b_n = \Omega(a_n)$ and $b_n \not\asymp a_n$.

### 1.3 Paper Organization

The rest of this paper is organized as follows. In the next section, we briefly introduce the elliptical distribution and review the marginal and multivariate Kendall's tau statistics. In Section 3, in the non-sparse setting, we propose the ECA method and study its theoretical performance. In Section 4, in the sparse setting, we propose a sparse ECA method via a combinatoric program and study its theoretical performance. A computationally efficient algorithm for conducting sparse ECA is provided in Section 5. Experiments on both synthetic and brain imaging data are provided in Section 6. More simulation results and all technical proofs are relegated to the supplementary materials.

## 2 Background

This section briefly reviews the elliptical distribution, and marginal and multivariate Kendall's tau statistics. In the sequel, we denote $\boldsymbol{X} \overset{\mathsf{d}}{=} \boldsymbol{Y}$ if random vectors $\boldsymbol{X}$ and $\boldsymbol{Y}$ have the same distribution.

### 2.1 Elliptical Distribution

The elliptical distribution is defined as follows. Let $\boldsymbol{\mu} \in \mathbb{R}^d$ and $\boldsymbol{\Sigma} \in \mathbb{R}^{d\times d}$ with $\mathrm{rank}(\boldsymbol{\Sigma}) = q \le d$. A $d$-dimensional random vector $\boldsymbol{X}$ has an elliptical distribution, denoted by $\boldsymbol{X} \sim EC_d(\boldsymbol{\mu}, \boldsymbol{\Sigma}, \xi)$, if it has a stochastic representation

$$\boldsymbol{X} \overset{\mathsf{d}}{=} \boldsymbol{\mu} + \xi \mathbf{A}\boldsymbol{U}, \tag{2.1}$$

where $\boldsymbol{U}$ is a uniform random vector on the unit sphere in $\mathbb{R}^q$, $\xi \geq 0$ is a scalar random variable independent of $\boldsymbol{U}$, and $\mathbf{A} \in \mathbb{R}^{d\times q}$ is a deterministic matrix satisfying $\mathbf{A}\mathbf{A}^T = \mathbf{\Sigma}$. Here $\mathbf{\Sigma}$ is called the scatter matrix. In this paper, we only consider continuous elliptical distributions with $\mathbb{P}(\xi = 0) = 0$.

An equivalent definition of the elliptical distribution is through the characteristic function $\exp(\mathrm{i}\boldsymbol{t}^T\boldsymbol{\mu})\psi(\boldsymbol{t}^T\mathbf{\Sigma}\boldsymbol{t})$, where $\psi$ is a properly defined characteristic function and $\mathrm{i} := \sqrt{-1}$. $\xi$ and $\psi$ are mutually determined. In this setting, we denote by $\boldsymbol{X} \sim EC_d(\boldsymbol{\mu}, \mathbf{\Sigma}, \psi)$. The elliptical family is closed under independent sums, and the marginal and conditional distributions of an elliptical distribution are also elliptically distributed.

Compared to the Gaussian family, the elliptical family provides more flexibility in modeling complex data. First, the elliptical family can model heavy-tail distributions (in contrast, Gaussian is light-tailed with exponential tail bounds). Secondly, the elliptical family can be used to model nontrivial tail dependence between variables (Hult and Lindskog, 2002), i.e., different variables tend to go to extremes together (in contrast, Gaussian family can not capture any tail dependence). The capability to handle heavy-tailed distributions and tail dependence is important for modeling many datasets, including: (1) financial data (almost all the financial data are heavy-tailed with nontrivial tail dependence (Rachev, 2003; Čižek et al., 2005)); (2) genomics data (Liu et al., 2003; Posekany et al., 2011); (3) bioimaging data (Ruttimann et al., 1998).

In the sequel, we assume that $\mathbb{E}\xi^2 < \infty$ so that the covariance matrix $\mathrm{Cov}(\boldsymbol{X})$ is well defined. For model identifiability, we further assume that $\mathbb{E}\xi^2 = q$ so that $\mathrm{Cov}(\boldsymbol{X}) = \mathbf{\Sigma}$. Of note, the assumption that the covariance matrix of $\boldsymbol{X}$ exists is added only for presentation clearness. In particular, the follow-up theorems still hold without any requirement on the moments of $\xi$. Actually, ECA still works even when $\mathbb{E}\xi = \infty$ due to its construction (see, for example, Equation (2.6) and follow-up discussions).

## 2.2 Marginal Rank-Based Estimators

In this section we briefly review the marginal rank-based estimator using the Kendall's tau statistic. This statistic plays a vital role in estimating the leading eigenvectors of the generalized correlation matrix $\mathbf{\Sigma}^0$ in Han and Liu (2014). Letting $\boldsymbol{X} := (X_1, \ldots, X_d)^T \in \mathbb{R}^d$ with $\widetilde{\boldsymbol{X}} := (\widetilde{X}_1, \ldots, \widetilde{X}_d)^T$ an independent copy of $\boldsymbol{X}$, the population Kendall's tau statistic is defined as:

$$\tau(X_j, X_k) := \mathrm{Cov}(\mathrm{sign}(X_j - \widetilde{X}_j), \mathrm{sign}(X_k - \widetilde{X}_k)).$$

Let $\boldsymbol{X}_1, \ldots, \boldsymbol{X}_n \in \mathbb{R}^d$ with $\boldsymbol{X}_i := (X_{i1}, \ldots, X_{id})^T$ be $n$ independent observations of $\boldsymbol{X}$. The sample Kendall's tau statistic is defined as:

$$\widehat{\tau}_{jk}(\boldsymbol{X}_1, \ldots, \boldsymbol{X}_n) := \frac{2}{n(n-1)} \sum_{1\leq i<i'\leq n} \mathrm{sign}(X_{ij} - X_{i'j})\mathrm{sign}(X_{ik} - X_{i'k}).$$

It is easy to verify that $\mathbb{E}\widehat{\tau}_{jk}(\boldsymbol{X}_1, \ldots, \boldsymbol{X}_n) = \tau(X_j, X_k)$. Let $\widehat{\mathbf{R}} = [\widehat{\mathbf{R}}_{jk}] \in \mathbb{R}^{d\times d}$, with $\widehat{\mathbf{R}}_{jk} = \sin(\frac{\pi}{2}\widehat{\tau}_{jk}(\boldsymbol{X}_1, \ldots, \boldsymbol{X}_n))$, be the Kendall's tau correlation matrix. The marginal rank-based estimator $\widetilde{\boldsymbol{\theta}}_1$ used by TCA is obtained by plugging $\widehat{\mathbf{R}}$ into the optimization formulation in Vu and Lei (2012). When $\boldsymbol{X} \sim EC_d(\boldsymbol{\mu}, \boldsymbol{\Sigma}, \xi)$ and under mild conditions, Han and Liu (2014) showed that

$$\mathbb{E}|\sin\angle(\widetilde{\boldsymbol{\theta}}_1, \boldsymbol{u}_1(\boldsymbol{\Sigma}^0))| = O\left(s\sqrt{\frac{\log d}{n}}\right),$$

where $s := \|\boldsymbol{u}_1(\boldsymbol{\Sigma}^0)\|_0$ and $\boldsymbol{\Sigma}^0$ is the generalized correlation matrix of $\boldsymbol{X}$. However, TCA is a variant of the scale-invariant PCA and can only estimate the leading eigenvectors of the correlation matrix. How, then, to estimate the leading eigenvector of the covariance matrix in high dimensional elliptical models? A straightforward approach is to exploit a covariance matrix estimator $\widehat{\mathbf{S}} := [\widehat{\mathbf{S}}_{jk}]$, defined as

$$\widehat{\mathbf{S}}_{jk} = \widehat{\mathbf{R}}_{jk} \cdot \widehat{\sigma}_j \widehat{\sigma}_k, \tag{2.2}$$

where $\{\widehat{\sigma}_j\}_{j=1}^d$ are sample standard deviations. However, since the elliptical distribution can be heavy-tailed, estimating the standard deviations is challenging and requires strong moment conditions. In this paper, we solve this problem by resorting to the multivariate rank-based method.

### 2.3 Multivariate Kendall's tau

Let $\boldsymbol{X} \sim EC_d(\boldsymbol{\mu}, \boldsymbol{\Sigma}, \xi)$ and $\widetilde{\boldsymbol{X}}$ be an independent copy of $\boldsymbol{X}$. The population multivariate Kendall's tau matrix, denoted by $\mathbf{K} \in \mathbb{R}^{d \times d}$, is defined as:

$$\mathbf{K} := \mathbb{E}\left(\frac{(\boldsymbol{X} - \widetilde{\boldsymbol{X}})(\boldsymbol{X} - \widetilde{\boldsymbol{X}})^T}{\|\boldsymbol{X} - \widetilde{\boldsymbol{X}}\|_2^2}\right). \tag{2.3}$$

Let $\boldsymbol{X}_1, \ldots, \boldsymbol{X}_n \in \mathbb{R}^d$ be $n$ independent data points of a random vector $\boldsymbol{X} \sim EC_d(\boldsymbol{\mu}, \boldsymbol{\Sigma}, \xi)$. The definition of the multivariate Kendall's tau in (2.3) motivates the following sample version multivariate Kendall's tau estimator, which is a second-order U-statistic:

$$\widehat{\mathbf{K}} := \frac{2}{n(n-1)} \sum_{i'<i} \frac{(\boldsymbol{X}_i - \boldsymbol{X}_{i'})(\boldsymbol{X}_i - \boldsymbol{X}_{i'})^T}{\|\boldsymbol{X}_i - \boldsymbol{X}_{i'}\|_2^2}. \tag{2.4}$$

It is obvious that $\mathbb{E}(\widehat{\mathbf{K}}) = \mathbf{K}$, and both $\mathbf{K}$ and $\widehat{\mathbf{K}}$ are positive semidefinite (PSD) matrices of trace 1. Moreover, the kernel of the U-statistic $k_{\mathsf{MK}}(\cdot) : \mathbb{R}^d \times \mathbb{R}^d \to \mathbb{R}^{d \times d}$,

$$k_{\mathsf{MK}}(\boldsymbol{X}_i, \boldsymbol{X}_{i'}) := \frac{(\boldsymbol{X}_i - \boldsymbol{X}_{i'})(\boldsymbol{X}_i - \boldsymbol{X}_{i'})^T}{\|\boldsymbol{X}_i - \boldsymbol{X}_{i'}\|_2^2}, \tag{2.5}$$

is bounded under the spectral norm, i.e., $\|k_{\mathsf{MK}}(\cdot)\|_2 \leq 1$. Intuitively, such a boundedness property makes the U-statistic $\widehat{\mathbf{K}}$ more amenable to theoretical analysis. Moreover, it is worth noting that $k_{\mathsf{MK}}(\boldsymbol{X}_i, \boldsymbol{X}_{i'})$ is a distribution-free kernel, i.e., for any continuous $\boldsymbol{X} \sim EC_d(\boldsymbol{\mu}, \boldsymbol{\Sigma}, \xi)$ with the generating variable $\xi$,

$$k_{\mathsf{MK}}(\boldsymbol{X}_i, \boldsymbol{X}_{i'}) \stackrel{\mathsf{d}}{=} k_{\mathsf{MK}}(\boldsymbol{Z}_i, \boldsymbol{Z}_{i'}), \tag{2.6}$$

where $\boldsymbol{Z}_i$ and $\boldsymbol{Z}_{i'}$ follow $\boldsymbol{Z} \sim N_d(\boldsymbol{\mu}, \boldsymbol{\Sigma})$. This can be proved using the closedness of the elliptical family under independent sums and the property that $\boldsymbol{Z}$ is a stochastic scaling of $\boldsymbol{X}$ (check Lemma B.7 for details). Accordingly, as will be shown later, the convergence of $\widehat{\mathbf{K}}$ to $\mathbf{K}$ does not depend on the generating variable $\xi$, and hence $\widehat{\mathbf{K}}$ enjoys the same distribution-free property as the Tyler's M estimator (Tyler, 1987). However, the multivariate Kendall's tau can be directly extended to analyze high dimensional data, while the Tyler's M estimator cannot[1].

The multivariate Kendall's tau can be viewed as the covariance matrix of the self-normalized data $\{(\boldsymbol{X}_i - \boldsymbol{X}_{i'})/\|\boldsymbol{X}_i - \boldsymbol{X}_{i'}\|_2\}_{i>i'}$. It is immediate to see that $\mathbf{K}$ is not identical or proportional to the covariance matrix $\boldsymbol{\Sigma}$ of $\boldsymbol{X}$. However, the following proposition, essentially coming from Marden (1999) and Croux et al. (2002) (also explicitly stated as Theorem 4.4 in Oja (2010)), states that the eigenspace of the multivariate Kendall's tau statistic $\mathbf{K}$ is identical to the eigenspace of the covariance matrix $\boldsymbol{\Sigma}$. Its proof is given in the supplementary materials for completeness.

**Proposition 2.1.** Let $\boldsymbol{X} \sim EC_d(\boldsymbol{\mu}, \boldsymbol{\Sigma}, \xi)$ be a continuous distribution and $\mathbf{K}$ be the population multivariate Kendall's tau statistic. Then if $\operatorname{rank}(\boldsymbol{\Sigma}) = q$, we have

$$\lambda_j(\mathbf{K}) = \mathbb{E}\left(\frac{\lambda_j(\boldsymbol{\Sigma})Y_j^2}{\lambda_1(\boldsymbol{\Sigma})Y_1^2 + \ldots + \lambda_q(\boldsymbol{\Sigma})Y_q^2}\right), \tag{2.7}$$

where $\boldsymbol{Y} := (Y_1, \ldots, Y_q)^T \sim N_q(\mathbf{0}, \mathbf{I}_q)$ is a standard multivariate Gaussian distribution. In addition, $\mathbf{K}$ and $\boldsymbol{\Sigma}$ share the same eigenspace with the same descending order of the eigenvalues.

Proposition 2.1 shows that, to recover the eigenspace of the covariance matrix $\boldsymbol{\Sigma}$, we can resort to recovering the eigenspace of $\mathbf{K}$, which, as is discussed above, can be more efficiently estimated using $\widehat{\mathbf{K}}$.

**Remark 2.2.** Proposition 2.1 shows that the eigenspaces of $\mathbf{K}$ and $\boldsymbol{\Sigma}$ are identical and the eigenvalues of $\mathbf{K}$ only depend on the eigenvalues of $\boldsymbol{\Sigma}$. Therefore, if we can theoretically calculate the relationships between $\{\lambda_j(\mathbf{K})\}_{j=1}^d$ and $\{\lambda_j(\boldsymbol{\Sigma})\}_{j=1}^d$, we can recover $\boldsymbol{\Sigma}$ using $\widehat{\mathbf{K}}$. When, for example, $\lambda_1(\boldsymbol{\Sigma}) = \cdots = \lambda_q(\boldsymbol{\Sigma})$, this relationship is calculable. In particular, it can be shown (check, for example, Section 3 in Bilodeau and Brenner (1999)) that

$$\frac{Y_j^2}{Y_1^2 + \cdots + Y_q^2} \sim \operatorname{Beta}\left(\frac{1}{2}, \frac{q-1}{2}\right), \quad \text{for } j = 1, \ldots, q,$$

where $\operatorname{Beta}(\alpha, \beta)$ is the beta distribution with parameters $\alpha$ and $\beta$. Accordingly, $\lambda_j(\mathbf{K}) = \mathbb{E}(Y_j^2/(Y_1^2 + \cdots + Y_q^2)) = 1/q$. The general relationship between $\{\lambda_j(\mathbf{K})\}_{j=1}^d$ and $\{\lambda_j(\boldsymbol{\Sigma})\}_{j=1}^d$ is non-linear. For example, when $d = 2$, Croux et al. (2002) showed that

$$\lambda_j(\mathbf{K}) = \frac{\sqrt{\lambda_j(\boldsymbol{\Sigma})}}{\sqrt{\lambda_1(\boldsymbol{\Sigma})} + \sqrt{\lambda_2(\boldsymbol{\Sigma})}}, \quad \text{for } j = 1, 2.$$

[1]The Tyler's M estimator cannot be directly applied to study high dimensional data because of both theoretical and empirical reasons. Theoretically, to the authors' knowledge, the sharpest sufficient condition to guarantee its consistency still requires $d = o(n^{1/2})$ (Duembgen, 1997). Empirically, our simulations show that Tyler's M estimator always fails to converge when $d > n$.

## 3 ECA: Non-Sparse Setting

In this section we propose and study the ECA method in the non-sparse setting when $\lambda_1(\mathbf{\Sigma})$ is distinct. In particular, we do not assume sparsity of $\boldsymbol{u}_1(\mathbf{\Sigma})$. Without the sparsity assumption, we propose to use the leading eigenvector $\boldsymbol{u}_1(\widehat{\mathbf{K}})$ to estimate $\boldsymbol{u}_1(\mathbf{K}) = \boldsymbol{u}_1(\mathbf{\Sigma})$:

$$\text{The ECA estimator}: \boldsymbol{u}_1(\widehat{\mathbf{K}}) \text{ (the leading eigenvector of } \widehat{\mathbf{K}}),$$

where $\widehat{\mathbf{K}}$ is defined in (2.4). For notational simplicity, in the sequel we assume that the sample size $n$ is even. When $n$ is odd, we can always use $n-1$ data points without affecting the obtained rate of convergence.

The approximation error of $\boldsymbol{u}_1(\widehat{\mathbf{K}})$ to $\boldsymbol{u}_1(\mathbf{K})$ is related to the convergence of $\widehat{\mathbf{K}}$ to $\mathbf{K}$ under the spectral norm via the Davis-Kahan inequality (Davis and Kahan, 1970; Wedin, 1972). In detail, for any two vectors $\boldsymbol{v}_1, \boldsymbol{v}_2 \in \mathbb{R}^d$, let $\sin\angle(\boldsymbol{v}_1, \boldsymbol{v}_2)$ be the sine of the angle between $\boldsymbol{v}_1$ and $\boldsymbol{v}_2$, with

$$|\sin\angle(\boldsymbol{v}_1, \boldsymbol{v}_2)| := \sqrt{1 - (\boldsymbol{v}_1^T \boldsymbol{v}_2)^2}.$$

The Davis-Kahan inequality states that the approximation error of $\boldsymbol{u}_1(\widehat{\mathbf{K}})$ to $\boldsymbol{u}_1(\mathbf{K})$ is controlled by $\|\widehat{\mathbf{K}} - \mathbf{K}\|_2$ divided by the eigengap between $\lambda_1(\mathbf{K})$ and $\lambda_2(\mathbf{K})$:

$$|\sin\angle(\boldsymbol{u}_1(\widehat{\mathbf{K}}), \boldsymbol{u}_1(\mathbf{K}))| \le \frac{2}{\lambda_1(\mathbf{K}) - \lambda_2(\mathbf{K})} \|\widehat{\mathbf{K}} - \mathbf{K}\|_2. \tag{3.1}$$

Accordingly, to analyze the convergence rate of $\boldsymbol{u}_1(\widehat{\mathbf{K}})$ to $\boldsymbol{u}_1(\mathbf{K})$, we can focus on the convergence rate of $\widehat{\mathbf{K}}$ to $\mathbf{K}$ under the spectral norm. The next theorem shows that, for the elliptical distribution family, the convergence rate of $\widehat{\mathbf{K}}$ to $\mathbf{K}$ under the spectral norm is $\|\mathbf{K}\|_2 \sqrt{r^*(\mathbf{K}) \log d / n}$, where $r^*(\mathbf{K}) = \mathrm{Tr}(\mathbf{K})/\lambda_1(\mathbf{K})$ is the effective rank of $\mathbf{K}$ and must be less than or equal to $d$.

**Theorem 3.1.** Let $\boldsymbol{X}_1, \ldots, \boldsymbol{X}_n$ be $n$ independent observations of $\boldsymbol{X} \sim EC_d(\boldsymbol{\mu}, \mathbf{\Sigma}, \xi)$. Let $\widehat{\mathbf{K}}$ be the sample version of the multivariate Kendall's tau statistic defined in Equation (2.4). We have, provided that $n$ is sufficiently large such that

$$n \ge \frac{16}{3} \cdot (r^*(\mathbf{K}) + 1)(\log d + \log(1/\alpha)), \tag{3.2}$$

with probability larger than $1 - \alpha$,

$$\|\widehat{\mathbf{K}} - \mathbf{K}\|_2 \le \|\mathbf{K}\|_2 \sqrt{\frac{16}{3} \cdot \frac{(r^*(\mathbf{K}) + 1)(\log d + \log(1/\alpha))}{n}}.$$

**Remark 3.2.** The scaling requirement on $n$ in (3.2) is posed largely for presentation clearness. As a matter of fact, we could withdraw this scaling condition by posing a different upper bound on $\|\widehat{\mathbf{K}} - \mathbf{K}\|_2$. This is via employing a similar argument as in Wegkamp and Zhao (2016). However, we note such a scaling requirement is necessary for proving ECA consistency in our analysis, and was also enforced in the related literature (see, for example, Theorem 3.1 in Han and Liu (2016)).

There is a vast literature on bounding the spectral norm of a random matrix (see, for example, Vershynin (2010) and the references therein) and our proof relies on the matrix Bernstein inequality proposed in Tropp (2012), with a generalization to U-statistics following similar arguments as in Wegkamp and Zhao (2016) and Han and Liu (2016). We defer the proof to Section B.2.

Combining (3.1) and Theorem 3.1, we immediately have the following corollary, which characterizes the explicit rate of convergence for $|\sin\angle(\boldsymbol{u}_1(\widehat{\mathbf{K}}), \boldsymbol{u}_1(\mathbf{K})|$.

**Corollary 3.1.** Under the conditions of Theorem 3.1, provided that $n$ is sufficiently large such that

$$n \geq \frac{16}{3} \cdot (r^*(\mathbf{K}) + 1)(\log d + \log(1/\alpha)),$$

we have, with probability larger than $1 - \alpha$,

$$|\sin\angle(\boldsymbol{u}_1(\widehat{\mathbf{K}}), \boldsymbol{u}_1(\mathbf{K}))| \leq \frac{2\lambda_1(\mathbf{K})}{\lambda_1(\mathbf{K}) - \lambda_2(\mathbf{K})} \sqrt{\frac{16}{3} \cdot \frac{(r^*(\mathbf{K})+1)(\log d + \log(1/\alpha))}{n}}.$$

**Remark 3.3.** Corollary 3.1 indicates that it is not necessary to require $d/n \to 0$ for $\boldsymbol{u}_1(\widehat{\mathbf{K}})$ to be a consistent estimator of $\boldsymbol{u}_1(\mathbf{K})$. For example, when $\lambda_2(\mathbf{K})/\lambda_1(\mathbf{K})$ is upper bounded by an absolute constant strictly smaller than 1, $r^*(\mathbf{K}) \log d/n \to 0$ is sufficient to make $\boldsymbol{u}_1(\widehat{\mathbf{K}})$ a consistent estimator of $\boldsymbol{u}_1(\mathbf{K})$. Such an observation is consistent with the observations in the PCA theory (Lounici, 2014; Bunea and Xiao, 2015). On the other hand, Theorem 4.1 in the next section provides a rate of convergence $O_P(\lambda_1(\mathbf{K})\sqrt{d/n})$ for $\|\widehat{\mathbf{K}} - \mathbf{K}\|_2$. Therefore, the final rate of convergence for ECA, under various settings, can be expressed as $O_P(\sqrt{r^*(\mathbf{K}) \log d/n} \wedge \sqrt{d/n})$.

**Remark 3.4.** We note that Theorem 3.1 can also help to quantify the subspace estimation error via a variation of the Davis-Kahan inequality. In particular, let $\mathcal{P}^m(\widehat{\mathbf{K}})$ and $\mathcal{P}^m(\mathbf{K})$ be the projection matrices onto the span of $m$ leading eigenvectors of $\widehat{\mathbf{K}}$ and $\mathbf{K}$. Using Lemma 4.2 in Vu and Lei (2013), we have

$$\|\mathcal{P}^m(\widehat{\mathbf{K}}) - \mathcal{P}^m(\mathbf{K})\|_{\mathsf{F}} \leq \frac{2\sqrt{2m}}{\lambda_m(\mathbf{K}) - \lambda_{m+1}(\mathbf{K})} \|\widehat{\mathbf{K}} - \mathbf{K}\|_2, \tag{3.3}$$

so that $\|\mathcal{P}^m(\widehat{\mathbf{K}}) - \mathcal{P}^m(\mathbf{K})\|_{\mathsf{F}}$ can be controlled via a similar argument as in Corollary 3.1.

The above bounds are all related to the eigenvalues of $\mathbf{K}$. The next theorem connects the eigenvalues of $\mathbf{K}$ to the eigenvalues of $\boldsymbol{\Sigma}$, so that we can directly bound $\|\widehat{\mathbf{K}} - \mathbf{K}\|_2$ and $|\sin\angle(\boldsymbol{u}_1(\widehat{\mathbf{K}}), \boldsymbol{u}_1(\mathbf{K}))|$ using $\boldsymbol{\Sigma}$. In the sequel, let's denote $r^{**}(\boldsymbol{\Sigma}) := \|\boldsymbol{\Sigma}\|_{\mathsf{F}}/\lambda_1(\boldsymbol{\Sigma}) \leq \sqrt{d}$ to be the "second-order" effective rank of the matrix $\boldsymbol{\Sigma}$.

**Theorem 3.5** (The upper and lower bounds of $\lambda_j(\mathbf{K})$)**.** Letting $\boldsymbol{X} \sim EC_d(\boldsymbol{\mu}, \boldsymbol{\Sigma}, \xi)$, we have

$$\lambda_j(\mathbf{K}) \geq \frac{\lambda_j(\boldsymbol{\Sigma})}{\mathrm{Tr}(\boldsymbol{\Sigma}) + 4\|\boldsymbol{\Sigma}\|_{\mathsf{F}}\sqrt{\log d} + 8\|\boldsymbol{\Sigma}\|_2 \log d}\left(1 - \frac{\sqrt{3}}{d^2}\right),$$

and when $\mathrm{Tr}(\boldsymbol{\Sigma}) > 4\|\boldsymbol{\Sigma}\|_{\mathsf{F}}\sqrt{\log d}$,

$$\lambda_j(\mathbf{K}) \leq \frac{\lambda_j(\boldsymbol{\Sigma})}{\mathrm{Tr}(\boldsymbol{\Sigma}) - 4\|\boldsymbol{\Sigma}\|_{\mathsf{F}}\sqrt{\log d}} + \frac{1}{d^4}.$$

Using Theorem 3.5 and recalling that the trace of $\mathbf{K}$ is always 1, we can replace $r^*(\mathbf{K})$ by $(r^*(\boldsymbol{\Sigma})+4r^{**}(\boldsymbol{\Sigma})\sqrt{\log d}+8\log d)\cdot(1-\sqrt{3}d^{-2})^{-1}$ in Theorem 3.1. We also note that Theorem 3.5 can help understand the scaling of $\lambda_j(\mathbf{K})$ with regard to $\lambda_j(\boldsymbol{\Sigma})$. Actually, when $\|\boldsymbol{\Sigma}\|_{\mathsf{F}}\log d = \mathrm{Tr}(\boldsymbol{\Sigma})\cdot o(1)$, we have $\lambda_j(\mathbf{K}) \asymp \lambda_j(\boldsymbol{\Sigma})/\mathrm{Tr}(\boldsymbol{\Sigma})$, and accordingly, we can continue to write

$$\frac{\lambda_1(\mathbf{K})}{\lambda_1(\mathbf{K})-\lambda_2(\mathbf{K})} \asymp \frac{\lambda_1(\boldsymbol{\Sigma})}{\lambda_1(\boldsymbol{\Sigma})-\lambda_2(\boldsymbol{\Sigma})}.$$

In practice, $\|\boldsymbol{\Sigma}\|_{\mathsf{F}}\log d = \mathrm{Tr}(\boldsymbol{\Sigma})\cdot o(1)$ is a mild condition. For example, when the condition number of $\boldsymbol{\Sigma}$ is upper bounded by an absolute constant, we have $\mathrm{Tr}(\boldsymbol{\Sigma}) \asymp \|\boldsymbol{\Sigma}\|_{\mathsf{F}}\cdot\sqrt{d}$.

For later purpose (see, for example, Theorem 5.3), sometimes we also need to connect the elementwise maximum norm $\|\mathbf{K}\|_{\max}$ to that of $\|\boldsymbol{\Sigma}\|_{\max}$. The next corollary gives such a connection.

**Corollary 3.2** (The upper and lower bounds of $\|\mathbf{K}\|_{\max}$)**.** Letting $\boldsymbol{X} \sim EC_d(\boldsymbol{\mu},\boldsymbol{\Sigma},\xi)$, we have

$$\|\mathbf{K}\|_{\max} \geq \frac{\|\boldsymbol{\Sigma}\|_{\max}}{\mathrm{Tr}(\boldsymbol{\Sigma})+4\|\boldsymbol{\Sigma}\|_{\mathsf{F}}\sqrt{\log d}+8\|\boldsymbol{\Sigma}\|_2\log d}\left(1-\frac{\sqrt{3}}{d^2}\right),$$

and when $\mathrm{Tr}(\boldsymbol{\Sigma}) > 4\|\boldsymbol{\Sigma}\|_{\mathsf{F}}\sqrt{\log d}$,

$$\|\mathbf{K}\|_{\max} \leq \frac{\|\boldsymbol{\Sigma}\|_{\max}}{\mathrm{Tr}(\boldsymbol{\Sigma})-4\|\boldsymbol{\Sigma}\|_{\mathsf{F}}\sqrt{\log d}}+\frac{1}{d^4}.$$

# 4 Sparse ECA via a Combinatoric Program

We analyze the theoretical properties of ECA in the sparse setting, where we assume $\lambda_1(\boldsymbol{\Sigma})$ is distinct and $\|\boldsymbol{u}_1(\boldsymbol{\Sigma})\|_0 \leq s < d\wedge n$. In this section we study the ECA method using a combinatoric program. For any matrix $\mathbf{M}\in\mathbb{R}^{d\times d}$, we define the best $s$-sparse vector approximating $\boldsymbol{u}_1(\mathbf{M})$ as

$$\boldsymbol{u}_{1,s}(\mathbf{M}) := \mathop{\arg\max}_{\|\boldsymbol{v}\|_0\leq s,\|\boldsymbol{v}\|_2\leq 1} |\boldsymbol{v}^T\mathbf{M}\boldsymbol{v}|. \tag{4.1}$$

We propose to estimate $\boldsymbol{u}_1(\boldsymbol{\Sigma}) = \boldsymbol{u}_1(\mathbf{K})$ via a combinatoric program:

$$\text{Sparse ECA estimator via a combinatoric program} : \boldsymbol{u}_{1,s}(\widehat{\mathbf{K}}),$$

where $\widehat{\mathbf{K}}$ is defined in (2.4). Under the sparse setting, by definition we have $\boldsymbol{u}_{1,s}(\mathbf{K}) = \boldsymbol{u}_1(\mathbf{K}) = \boldsymbol{u}_1(\boldsymbol{\Sigma})$. On the other hand, $\boldsymbol{u}_{1,s}(\widehat{\mathbf{K}})$ can be calculated via a combinatoric program by exhaustively searching over all $s$ by $s$ submatrices of $\widehat{\mathbf{K}}$. This global search is not computationally efficient. However, the result in quantifying the approximation error of $\boldsymbol{u}_{1,s}(\widehat{\mathbf{K}})$ to $\boldsymbol{u}_1(\mathbf{K})$ is of strong theoretical interest. Similar algorithms were also studied in Vu and Lei (2012), Lounici (2013), Vu and Lei (2013), and Cai et al. (2015). Moreover, as will be seen in the next section, this will help clarify that a computationally efficient sparse ECA algorithm can attain the same convergence rate, though under a suboptimal scaling of $(n,d,s)$.

In the following we study the performance of $\boldsymbol{u}_{1,s}(\widehat{\mathbf{K}})$ in conducting sparse ECA. The approximation error of $\boldsymbol{u}_{1,s}(\widehat{\mathbf{K}})$ to $\boldsymbol{u}_1(\mathbf{K})$ is connected to the approximation error of $\widehat{\mathbf{K}}$ to $\mathbf{K}$ under the restricted spectral norm. This is due to the following Davis-Kahan type inequality provided in Vu and Lei (2012):

$$|\sin \angle(\boldsymbol{u}_{1,s}(\widehat{\mathbf{K}}), \boldsymbol{u}_{1,s}(\mathbf{K}))| \leq \frac{2}{\lambda_1(\mathbf{K}) - \lambda_2(\mathbf{K})} \|\widehat{\mathbf{K}} - \mathbf{K}\|_{2,2s}. \tag{4.2}$$

Accordingly, for studying $|\sin \angle(\boldsymbol{u}_{1,s}(\widehat{\mathbf{K}}), \boldsymbol{u}_{1,s}(\mathbf{K}))|$, we focus on studying the approximation error $\|\widehat{\mathbf{K}} - \mathbf{K}\|_{2,s}$. Before presenting the main results, we provide some extra notation. For any random variable $X \in \mathbb{R}$, we define the subgaussian ($\|\cdot\|_{\psi_2}$) and sub-exponential norms ($\|\cdot\|_{\psi_1}$) of $X$ as follows:

$$\|X\|_{\psi_2} := \sup_{k \geq 1} k^{-1/2} (\mathbb{E}|X|^k)^{1/k} \quad \text{and} \quad \|X\|_{\psi_1} := \sup_{k \geq 1} k^{-1} (\mathbb{E}|X|^k)^{1/k}. \tag{4.3}$$

Any $d$-dimensional random vector $\boldsymbol{X} \in \mathbb{R}^d$ is said to be subgaussian distributed with the subgaussian constant $\sigma$ if

$$\|\boldsymbol{v}^T \boldsymbol{X}\|_{\psi_2} \leq \sigma, \quad \text{for any } \boldsymbol{v} \in \mathbb{S}^{d-1}.$$

Moreover, we define the self-normalized operator $S(\cdot)$ for any random vector to be

$$S(\boldsymbol{X}) := (\boldsymbol{X} - \widetilde{\boldsymbol{X}})/\|\boldsymbol{X} - \widetilde{\boldsymbol{X}}\|_2 \quad \text{where } \widetilde{\boldsymbol{X}} \text{ is an independent copy of X.} \tag{4.4}$$

It is immediate that $\mathbf{K} = \mathbb{E} S(\boldsymbol{X}) S(\boldsymbol{X})^T$.

The next theorem provides a general result in quantifying the approximation error of $\widehat{\mathbf{K}}$ to $\mathbf{K}$ with regard to the restricted spectral norm.

**Theorem 4.1.** Let $\boldsymbol{X}_1, \ldots, \boldsymbol{X}_n$ be $n$ observations of $\boldsymbol{X} \sim EC_d(\boldsymbol{\mu}, \boldsymbol{\Sigma}, \xi)$. Let $\widehat{\mathbf{K}}$ be the sample version multivariate Kendall's tau statistic defined in Equation (2.4). We have, when $(s \log(ed/s) + \log(1/\alpha))/n \to 0$, for $n$ sufficiently large, with probability larger than $1 - 2\alpha$,

$$\|\widehat{\mathbf{K}} - \mathbf{K}\|_{2,s} \leq \left( \sup_{\boldsymbol{v} \in \mathbb{S}^{d-1}} 2\|\boldsymbol{v}^T S(\boldsymbol{X})\|_{\psi_2}^2 + \|\mathbf{K}\|_2 \right) \cdot C_0 \sqrt{\frac{s(3 + \log(d/s)) + \log(1/\alpha)}{n}},$$

for some absolute constant $C_0 > 0$. Here $\sup_{\boldsymbol{v} \in \mathbb{S}^{d-1}} \|\boldsymbol{v}^T \boldsymbol{X}\|_{\psi_2}$ can be further written as

$$\sup_{\boldsymbol{v} \in \mathbb{S}^{d-1}} \|\boldsymbol{v}^T S(\boldsymbol{X})\|_{\psi_2} = \sup_{\boldsymbol{v} \in \mathbb{S}^{d-1}} \left\| \frac{\sum_{i=1}^d v_i \lambda_i^{1/2}(\boldsymbol{\Sigma}) Y_i}{\sqrt{\sum_{i=1}^d \lambda_i(\boldsymbol{\Sigma}) Y_i^2}} \right\|_{\psi_2} \leq 1, \tag{4.5}$$

where $\boldsymbol{v} := (v_1, \ldots, v_d)^T$ and $(Y_1, \ldots, Y_d)^T \sim N_d(\mathbf{0}, \mathbf{I}_d)$.

It is obvious that $S(\boldsymbol{X})$ is subgaussian with variance proxy 1. However, typically, a sharper upper bound can be obtained. The next theorem shows, under various settings, the upper bound can be of the same order as $1/q$, which is much smaller than 1. Combined with Theorem 4.1, these results give an upper bound of $\|\widehat{\mathbf{K}} - \mathbf{K}\|_{2,s}$.

**Theorem 4.2.** Let $\boldsymbol{X}_1, \ldots, \boldsymbol{X}_n$ be $n$ observations of $\boldsymbol{X} \sim EC_d(\boldsymbol{\mu}, \boldsymbol{\Sigma}, \xi)$ with $\text{rank}(\boldsymbol{\Sigma}) = q$ and $\|\boldsymbol{u}_1(\boldsymbol{\Sigma})\|_0 \leq s$. Let $\widehat{\mathbf{K}}$ be the sample version multivariate Kendall's tau statistic defined in Equation (2.4). We have,

$$\sup_{\boldsymbol{v} \in \mathbb{S}^{d-1}} \|\boldsymbol{v}^T S(\boldsymbol{X})\|_{\psi_2} \leq \sqrt{\frac{\lambda_1(\boldsymbol{\Sigma})}{\lambda_q(\boldsymbol{\Sigma})} \cdot \frac{2}{q} \wedge 1},$$

and accordingly, when $(s \log(ed/s) + \log(1/\alpha))/n \to 0$, with probability at least $1 - 2\alpha$,

$$\|\widehat{\mathbf{K}} - \mathbf{K}\|_{2,s} \leq C_0 \left\{ \left( \frac{4\lambda_1(\boldsymbol{\Sigma})}{q\lambda_q(\boldsymbol{\Sigma})} \wedge 1 \right) + \lambda_1(\mathbf{K}) \right\} \sqrt{\frac{s(3 + \log(d/s)) + \log(1/\alpha)}{n}}.$$

Similar to Theorem 3.1, we wish to show that $\|\widehat{\mathbf{K}} - \mathbf{K}\|_{2,s} = O_P(\lambda_1(\mathbf{K})\sqrt{s \log(ed/s)/n})$. In the following, we provide several examples such that $\sup_{\boldsymbol{v}} \|\boldsymbol{v}^T S(\boldsymbol{X})\|_{\psi_2}^2$ is of the same order as $\lambda_1(\mathbf{K})$, so that, via Theorem 4.2, the desired rate is attained.

- **Condition number controlled:** Bickel and Levina (2008) considered the covariance matrix model where the condition number of the covariance matrix $\boldsymbol{\Sigma}$, $\lambda_1(\boldsymbol{\Sigma})/\lambda_d(\boldsymbol{\Sigma})$, is upper bounded by an absolute constant. Under this condition, we have

  $$\sup_{\boldsymbol{v}} \|\boldsymbol{v}^T S(\boldsymbol{X})\|_{\psi_2}^2 \asymp d^{-1},$$

  and applying Theorem 3.5 we also have

  $$\lambda_j(\mathbf{K}) \asymp \frac{\lambda_j(\boldsymbol{\Sigma})}{\text{Tr}(\boldsymbol{\Sigma})} \asymp d^{-1}.$$

  Accordingly, we conclude that $\sup_{\boldsymbol{v}} \|\boldsymbol{v}^T S(\boldsymbol{X})\|_{\psi_2}^2$ and $\lambda_1(\mathbf{K})$ are of the same order.

- **Spike covariance model:** Johnstone and Lu (2009) considered the following simple spike covariance model:

  $$\boldsymbol{\Sigma} = \beta \boldsymbol{v}\boldsymbol{v}^T + a^2 \mathbf{I}_d,$$

  where $\beta, a > 0$ are two positive real numbers and $\boldsymbol{v} \in \mathbb{S}^{d-1}$. In this case, we have, when $\beta = o(da^2/\sqrt{\log d})$ or $\beta = \Omega(da^2)$,

  $$\sup_{\boldsymbol{v}} \|\boldsymbol{v}^T S(\boldsymbol{X})\|_{\psi_2}^2 \asymp \frac{\beta + a^2}{da^2} \wedge 1 \quad \text{and} \quad \lambda_1(\mathbf{K}) \asymp \frac{\beta + a^2}{\beta + da^2}.$$

  A simple calculation shows that $\sup_{\boldsymbol{v}} \|\boldsymbol{v}^T S(\boldsymbol{X})\|_{\psi_2}^2$ and $\lambda_1(\mathbf{K})$ are of the same order.

- **Multi-Factor Model:** Fan et al. (2008) considered a multi-factor model, which is also related to the general spike covariance model (Ma, 2013):

  $$\boldsymbol{\Sigma} = \sum_{j=1}^{m} \beta_j \boldsymbol{v}_j \boldsymbol{v}_j^T + \boldsymbol{\Sigma}_u,$$

where we have $\beta_1 \geq \beta_2 \geq \cdots \geq \beta_m > 0$, $\boldsymbol{v}_1, \ldots, \boldsymbol{v}_m \in \mathbb{S}^{d-1}$ are orthogonal to each other, and $\boldsymbol{\Sigma}_u$ is a diagonal matrix. For simplicity, we assume that $\boldsymbol{\Sigma}_u = a^2 \mathbf{I}_d$. When $\sum \beta_j^2 = o(d^2 a^4 / \log d)$, we have

$$\sup_{\boldsymbol{v}} \|\boldsymbol{v}^T S(\boldsymbol{X})\|_{\psi_2}^2 \asymp \frac{\beta_1 + a^2}{da^2} \wedge 1 \quad \text{and} \quad \lambda_1(\mathbf{K}) \asymp \frac{\beta_1 + a^2}{\sum_{j=1}^m \beta_j + da^2},$$

and $\sup_{\boldsymbol{v}} \|\boldsymbol{v}^T S(\boldsymbol{X})\|_{\psi_2}^2$ and $\lambda_1(\mathbf{K})$ are of the same order if, for example, $\sum_{j=1}^m \beta_j = O(da^2)$.

Equation (4.2) and Theorem 4.2 together give the following corollary, which quantifies the convergence rate of the sparse ECA estimator calculated via the combinatoric program in (4.1).

**Corollary 4.1.** Under the condition of Theorem 4.2, if we have $(s \log(ed/s) + \log(1/\alpha))/n \to 0$, for $n$ sufficiently large, with probability larger than $1 - 2\alpha$,

$$|\sin \angle(\boldsymbol{u}_{1,s}(\widehat{\mathbf{K}}), \boldsymbol{u}_{1,s}(\mathbf{K}))| \leq \frac{2C_0(4\lambda_1(\boldsymbol{\Sigma})/q\lambda_q(\boldsymbol{\Sigma}) \wedge 1 + \lambda_1(\mathbf{K}))}{\lambda_1(\mathbf{K}) - \lambda_2(\mathbf{K})} \cdot \sqrt{\frac{2s(3 + \log(d/2s)) + \log(1/\alpha)}{n}}.$$

**Remark 4.3.** The restricted spectral norm convergence result obtained in Theorem 4.2 is also applicable to analyzing principal subspace estimation accuracy. Following the notation in Vu and Lei (2013), we define the principal subspace estimator to the space spanned by the top $m$ eigenvectors of any given matrix $\mathbf{M} \in \mathbb{R}^{d \times d}$ as

$$\mathbf{U}_{m,s}(\mathbf{M}) := \operatorname*{arg\,max}_{\mathbf{V} \in \mathbb{R}^{d \times m}} \left\langle \mathbf{M}, \mathbf{V}\mathbf{V}^T \right\rangle, \quad \text{subject to} \quad \sum_{j=1}^d \mathbb{I}(\mathbf{V}_{j*} \neq 0) \leq s, \tag{4.6}$$

where $V_{j*}$ is the $j$-th row of $\mathbf{M}$ and the indicator function returns 0 if and only if $\mathbf{V}_{j*} = \mathbf{0}$. We then have

$$\|\mathbf{U}_{m,s}(\widehat{\mathbf{K}})\mathbf{U}_{m,s}(\widehat{\mathbf{K}})^T - \mathbf{U}_{m,s}(\mathbf{K})\mathbf{U}_{m,s}(\mathbf{K})^T\|_{\mathsf{F}} \leq \frac{2\sqrt{2m}}{\lambda_m(\mathbf{K}) - \lambda_{m+1}(\mathbf{K})} \cdot \|\widehat{\mathbf{K}} - \mathbf{K}\|_{2,2ms}.$$

An explicit statement of the above inequality can be found in Wang et al. (2013).

# 5 Sparse ECA via a Computationally Efficient Program

There is a vast literature studying computationally efficient algorithms for estimating sparse $\boldsymbol{u}_1(\boldsymbol{\Sigma})$. In this section we focus on one such algorithm for conducting sparse ECA by combining the Fantope projection (Vu et al., 2013) with the truncated power method (Yuan and Zhang, 2013).

## 5.1 Fantope Projection

In this section we first review the algorithm and theory developed in Vu et al. (2013) for sparse subspace estimation, and then provide some new analysis in obtaining the sparse leading eigenvector estimators.

Let $\mathbf{\Pi}_m := \mathbf{V}_m\mathbf{V}_m^T$ where $\mathbf{V}_m$ is the combination of the $m$ leading eigenvectors of $\mathbf{K}$. It is well known that $\mathbf{\Pi}_m$ is the optimal rank-$m$ projection onto $\mathbf{K}$. Similarly as in (4.6), we define $s_\Pi$ to be the number of nonzero columns in $\mathbf{\Pi}_m$.

We then introduce the sparse principal subspace estimator $\mathbf{X}_m$ corresponding to the space spanned by the first $m$ leading eigenvectors of the multivariate Kendall's tau matrix $\widehat{\mathbf{K}}$. To induce sparsity, $\mathbf{X}_m$ is defined to be the solution to the following convex program:

$$\mathbf{X}_m := \mathop{\arg\max}_{\mathbf{M}\in\mathbb{R}^{d\times d}} \langle \widehat{\mathbf{K}}, \mathbf{M}\rangle - \lambda \sum_{j,k} |\mathbf{M}_{jk}|, \quad \text{subject to } \mathbf{0} \preceq \mathbf{M} \preceq \mathbf{I}_d \text{ and } \mathrm{Tr}(\mathbf{M}) = m, \tag{5.1}$$

where for any two matrices $\mathbf{A}, \mathbf{B} \in \mathbb{R}^{d\times d}$, $\mathbf{A} \preceq \mathbf{B}$ represents that $\mathbf{B} - \mathbf{A}$ is positive semidefinite. Here $\{\mathbf{M} : \mathbf{0} \preceq \mathbf{M} \preceq \mathbf{I}_d, \mathrm{Tr}(\mathbf{M}) = m\}$ is a convex set called the Fantope. We then have the following deterministic theorem to quantify the approximation error of $\mathbf{X}_m$ to $\mathbf{\Pi}_m$.

**Theorem 5.1** (Vu et al. (2013))**.** If the tuning parameter $\lambda$ in (5.1) satisfies that $\lambda \geq \|\widehat{\mathbf{K}} - \mathbf{K}\|_{\max}$, we have

$$\|\mathbf{X}_m - \mathbf{\Pi}_m\|_{\mathsf{F}} \leq \frac{4 s_\Pi \lambda}{\lambda_m(\mathbf{K}) - \lambda_{m+1}(\mathbf{K})},$$

where we remind that $s_\Pi$ is the number of nonzero columns in $\mathbf{\Pi}_m$.

It is easy to see that $\mathbf{X}_m$ is symmetric and the rank of $\mathbf{X}_m$ must be greater than or equal to $m$, but is not necessarily exactly $m$. However, in various cases, dimension reduction for example, it is desired to estimate the top $m$ leading eigenvectors of $\mathbf{\Sigma}$, or equivalently, to estimate an exactly rank $m$ projection matrix. Noticing that $\mathbf{X}_m$ is a real symmetric matrix, we propose to use the following estimate $\widehat{\mathbf{X}}_m \in \mathbb{R}^{d\times d}$:

$$\widehat{\mathbf{X}}_m := \sum_{j\leq m} \boldsymbol{u}_j(\mathbf{X}_m)[\boldsymbol{u}_j(\mathbf{X}_m)]^T. \tag{5.2}$$

We then have the next theorem, which quantifies the distance between $\widehat{\mathbf{X}}_m$ and $\mathbf{\Pi}_m$.

**Theorem 5.2.** If $\lambda \geq \|\widehat{\mathbf{K}} - \mathbf{K}\|_{\max}$, we have

$$\|\widehat{\mathbf{X}}_m - \mathbf{\Pi}_m\|_{\mathsf{F}} \leq 4\|\mathbf{X}_m - \mathbf{\Pi}_m\|_{\mathsf{F}} \leq \frac{16 s_\Pi \lambda}{\lambda_m(\mathbf{K}) - \lambda_{m+1}(\mathbf{K})}.$$

## 5.2 A Computationally Efficient Algorithm

In this section we propose a computationally efficient algorithm to conduct sparse ECA via combining the Fantope projection with the truncated power algorithm proposed in Yuan and Zhang (2013). We focus on estimating the leading eigenvector of $\mathbf{K}$ since the rest can be iteratively estimated using the deflation method (Mackey, 2008).

The main idea here is to exploit the Fantope projection for constructing a good initial parameter for the truncated power algorithm and then perform iterative thresholding as in Yuan and Zhang (2013). We call this the Fantope-truncated power algorithm, or FTPM, for abbreviation. Before proceeding to the main algorithm, we first introduce some extra notation. For any vector $\boldsymbol{v} \in \mathbb{R}^d$ and an index set $J \subset \{1, \ldots, d\}$, we define the truncation function $\mathrm{TRC}(\cdot, \cdot)$ to be

$$\mathrm{TRC}(\boldsymbol{v}, J) := \big(v_1 \cdot \mathbb{I}(1 \in J), \;\; \ldots, \;\; v_d \cdot \mathbb{I}(d \in J)\big)^T, \tag{5.3}$$

where $\mathbb{I}(\cdot)$ is the indicator function. The initial parameter $\boldsymbol{v}^{(0)}$, then, is the normalized vector consisting of the largest entries in $\boldsymbol{u}_1(\mathbf{X}_1)$, where $\mathbf{X}_1$ is calculated in (5.1):

$$\boldsymbol{v}^{(0)} = \boldsymbol{w}^0/\|\boldsymbol{w}^0\|_2, \text{ where } \boldsymbol{w}^0 = \mathrm{TRC}(\boldsymbol{u}_1(\mathbf{X}_1), J_\delta) \text{ and } J_\delta = \{j : |(\boldsymbol{u}_1(\mathbf{X}_1))_j| \geq \delta\}. \tag{5.4}$$

We have $\|\boldsymbol{v}^{(0)}\|_0 = \mathrm{supp}\{j : |(\boldsymbol{u}_1(\mathbf{X}_1))_j| > 0\}$. Algorithm 1 then provides the detailed FTPM algorithm and the final FTPM estimator is denoted as $\widehat{\boldsymbol{u}}_{1,k}^{\mathsf{FT}}$.

---

**Algorithm 1** The FTPM algorithm. Within each iteration, a new sparse vector $\boldsymbol{v}^{(t)}$ with $\|\boldsymbol{v}^{(t)}\|_0 \leq k$ is updated. The algorithm terminates when $\|\boldsymbol{v}^{(t)} - \boldsymbol{v}^{(t-1)}\|_2$ is less than a given threshold $\epsilon$.

---

**Algorithm:** $\widehat{\boldsymbol{u}}_{1,k}^{\mathsf{FT}}(\widehat{\mathbf{K}}) \leftarrow \mathsf{FTPM}(\widehat{\mathbf{K}}, k, \epsilon)$
**Initialize:** $\mathbf{X}_1$ calculated by (5.1) with $m = 1$, $\boldsymbol{v}^{(0)}$ is calculated using (5.4), and $t \leftarrow 0$
**Repeat:**
  $t \leftarrow t + 1$
  $\boldsymbol{X}_y \leftarrow \widehat{\mathbf{K}}\boldsymbol{v}^{(t-1)}$
  If $\|\boldsymbol{X}_t\|_0 \leq k$, then $\boldsymbol{v}^{(t)} = \boldsymbol{X}_t/\|\boldsymbol{X}_t\|_2$
  Else, let $A_t$ be the indices of the elements in $\boldsymbol{X}_t$ with the $k$ largest absolute values
  $\boldsymbol{v}^{(t)} = \mathrm{TRC}(\boldsymbol{X}_t, A_t)/\|\,\mathrm{TRC}(\boldsymbol{X}_t, A_t)\|_2$
**Until convergence:** $\|\boldsymbol{v}^{(t)} - \boldsymbol{v}^{(t-1)}\|_2 \leq \epsilon$
$\widehat{\boldsymbol{u}}_{1,k}^{\mathsf{FT}}(\widehat{\mathbf{K}}) \leftarrow \boldsymbol{v}^{(t)}$
**Output:** $\widehat{\boldsymbol{u}}_{1,k}^{\mathsf{FT}}(\widehat{\mathbf{K}})$

---

In the rest of this section we study the approximation accuracy of $\widehat{\boldsymbol{u}}_{1,k}^{\mathsf{FT}}$ to $\boldsymbol{u}_1(\mathbf{K})$. Via observing Theorems 5.1 and 5.2, it is immediate that the approximation accuracy of $\boldsymbol{u}_1(\mathbf{X}_1)$ is related to $\|\widehat{\mathbf{K}} - \mathbf{K}\|_{\max}$. The next theorem gives a nonasymptotic upper bound of $\|\widehat{\mathbf{K}} - \mathbf{K}\|_{\max}$, and accordingly, combined with Theorems 5.1 and 5.2, gives an upper bound on $|\sin\angle(\boldsymbol{u}_1(\mathbf{X}_1), \boldsymbol{u}_1(\mathbf{K}))|$.

**Theorem 5.3.** Let $\boldsymbol{X}_1, \ldots, \boldsymbol{X}_n$ be $n$ observations of $\boldsymbol{X} \sim EC_d(\boldsymbol{\mu}, \boldsymbol{\Sigma}, \xi)$ with $\mathrm{rank}(\boldsymbol{\Sigma}) = q$ and $\|\boldsymbol{u}_1(\boldsymbol{\Sigma})\|_0 \leq s$. Let $\widehat{\mathbf{K}}$ be the sample version multivariate Kendall's tau statistic defined in Equation (2.4). If $(\log d + \log(1/\alpha))/n \to 0$, we have there exists some positive absolute constant $C_1$ such that for sufficiently large $n$, with probability at least $1 - \alpha^2$,

$$\|\widehat{\mathbf{K}} - \mathbf{K}\|_{\max} \leq C_1\Big(\frac{8\lambda_1(\boldsymbol{\Sigma})}{q\lambda_q(\boldsymbol{\Sigma})} + \|\mathbf{K}\|_{\max}\Big)\sqrt{\frac{\log d + \log(1/\alpha)}{n}}.$$

Accordingly, if

$$\lambda \geq C_1\Big(\frac{8\lambda_1(\boldsymbol{\Sigma})}{q\lambda_q(\boldsymbol{\Sigma})} + \|\mathbf{K}\|_{\max}\Big)\sqrt{\frac{\log d + \log(1/\alpha)}{n}}, \tag{5.5}$$

we have, with probability at least $1-\alpha^2$,

$$|\sin\angle(\boldsymbol{u}_1(\mathbf{X}_1),\boldsymbol{u}_1(\mathbf{K}))| \le \frac{8\sqrt{2}s\lambda}{\lambda_1(\mathbf{K})-\lambda_2(\mathbf{K})}.$$

Theorem 5.3 builds sufficient conditions under which $\boldsymbol{u}_1(\mathbf{X}_1)$ is a consistent estimator of $\boldsymbol{u}_1(\mathbf{K})$. In multiple settings — the "condition number controlled", "spike covariance model", and "multi-factor model" settings considered in Section 4 for example — when $\lambda \asymp \lambda_1(\mathbf{K})\sqrt{\log d/n}$, we have $|\sin\angle(\boldsymbol{u}_1(\mathbf{X}_1),\boldsymbol{u}_1(\mathbf{K}))| = O_P(s\sqrt{\log d/n})$. This is summarized in the next corollary.

**Corollary 5.1.** Under the conditions of Theorem 5.3, if we further have $\lambda_1(\boldsymbol{\Sigma})/q\lambda_q(\boldsymbol{\Sigma}) = O(\lambda_1(\mathbf{K}))$, $\|\boldsymbol{\Sigma}\|_{\mathsf{F}}\log d = \mathrm{Tr}(\boldsymbol{\Sigma})\cdot o(1)$, $\lambda_2(\boldsymbol{\Sigma})/\lambda_1(\boldsymbol{\Sigma})$ is upper bounded by an absolute constant less than 1, and $\lambda \asymp \lambda_1(\mathbf{K})\sqrt{\log d/n}$, then

$$|\sin\angle(\boldsymbol{u}_1(\mathbf{X}_1),\boldsymbol{u}_1(\mathbf{K}))| = O_P\Big(s\sqrt{\frac{\log d}{n}}\Big).$$

Corollary 5.1 is a direct consequence of Theorem 5.3 and Theorem 3.5, and its proof is omitted. We then turn to study the estimation error of $\widehat{\boldsymbol{u}}^{\mathsf{FT}}_{1,k}(\widehat{\mathbf{K}})$. By examining Theorem 4 in Yuan and Zhang (2013), for theoretical guarantee of fast rate of convergence, it is enough to show that $(\boldsymbol{v}^{(0)})^T\boldsymbol{u}_1(\mathbf{K})$ is lower bounded by an absolute constant larger than zero. In the next theorem, we show that, under mild conditions, this is true with high probability, and accordingly we can exploit the result in Yuan and Zhang (2013) to show that $\widehat{\boldsymbol{u}}^{\mathsf{FT}}_{1,k}(\widehat{\mathbf{K}})$ attains the same optimal convergence rate as that of $\boldsymbol{u}_{1,s}(\widehat{\mathbf{K}})$.

**Theorem 5.4.** Under the conditions of Corollary 5.1, let $J_0 := \{j : |(\boldsymbol{u}_1(\mathbf{K}))_j| = \Omega^0(s\log d/\sqrt{n})\}$. Set $\delta$ in (5.4) to be $\delta = C_2 s(\log d)/\sqrt{n}$ for some positive absolute constant $C_2$. If $s\sqrt{\log d/n}\to 0$, and $\|(\boldsymbol{u}_1(\mathbf{K}))_{J_0}\|_2 \ge C_3 > 0$ is lower bounded by an absolute positive constant, then, with probability tending to 1, $\|\boldsymbol{v}^{(0)}\|_0 \le s$ and $|(\boldsymbol{v}^{(0)})^T\boldsymbol{u}_1(\mathbf{K})|$ is lower bounded by $C_3/2$. Accordingly under the condition of Theorem 4 in Yuan and Zhang (2013), for $k\ge s$, we have

$$|\sin\angle(\widehat{\boldsymbol{u}}^{\mathsf{FT}}_{1,k}(\widehat{\mathbf{K}}),\boldsymbol{u}_1(\mathbf{K}))| = O_P\Big(\sqrt{\frac{(k+s)\log d}{n}}\Big).$$

**Remark 5.5.** Although a similar second step truncation is performed, the assumption that the largest entries in $\boldsymbol{u}_1(\mathbf{K})$ satisfy $\|(\boldsymbol{u}_1(\mathbf{K}))_{J_0}\|_2 \ge C_3$ is much weaker than the assumption in Theorem 3.2 of Vu et al. (2013) because we allow a lot of entries in the leading eigenvector to be small and not detectable. This is permissible since our aim is parameter estimation instead of guaranteeing the model selection consistency.

**Remark 5.6.** In practice, we can adaptively select the tuning parameter $k$ in Algorithm 1. One possible way is to use the criterion of Yuan and Zhang (2013), selecting $k$ that maximizes $(\widehat{\boldsymbol{u}}^{\mathsf{FT}}_{1,k}(\widehat{\mathbf{K}}))^T\cdot\widehat{\mathbf{K}}_{\mathrm{val}}\cdot\widehat{\boldsymbol{u}}^{\mathsf{FT}}_{1,k}(\widehat{\mathbf{K}})$, where $\widehat{\mathbf{K}}_{\mathrm{val}}$ is an independent empirical multivariate Kendall's tau statistic based on a separated sample set of the data. Yuan and Zhang (2013) showed that such a heuristic performed quite well in applications.

**Remark 5.7.** In Corollary 5.1 and Theorem 5.4, we assume that $\lambda$ is in the same scale of $\lambda_1(\mathbf{K})\sqrt{\log d/n}$. In practice, $\lambda$ is a tuning parameter. Here we can select $\lambda$ using similar data driven estimation procedures as proposed in Lounici (2013) and Wegkamp and Zhao (2016). The main idea is to replace the population quantities with their corresponding empirical versions in (5.5). We conjecture that similar theoretical behaviors can be anticipated by the data driven way.

# 6 Numerical Experiments

In this section we use both synthetic and real data to investigate the empirical usefulness of ECA. We use the FTPM algorithm described in Algorithm 1 for parameter estimation. To estimate more than one leading eigenvectors, we exploit the deflation method proposed in Mackey (2008). Here the cardinalities of the support sets of the leading eigenvectors are treated as tuning parameters. The following three methods are considered:

- TP: Sparse PCA method on the Pearson's sample covariance matrix;
- TCA: Transelliptical component analysis based on the transformed Kendall's tau covariance matrix shown in Equation (2.2);
- ECA: Elliptical component analysis based on the multivariate kendall's tau matrix.

For fairness of comparison, TCA and TP also exploit the FTPM algorithm, while using the Kendall's tau covariance matrix and Pearson's sample covariance matrix as the input matrix. The tuning parameter $\lambda$ in (5.1) is selected using the method discussed in Remark 5.7, and the truncation value $\delta$ in (5.4) is selected such that $\|\boldsymbol{v}^{(0)}\|_0 = s$ for the pre-specified sparsity level $s$.

## 6.1 Simulation Study

In this section, we conduct a simulation study to back up the theoretical results and further investigate the empirical performance of ECA.

### 6.1.1 Dependence on Sample Size and Dimension

We first illustrate the dependence of the estimation accuracy of the sparse ECA estimator on the triplet $(n, d, s)$. We adopt the data generating schemes of Yuan and Zhang (2013) and Han and Liu (2014). More specifically, we first create a covariance matrix $\mathbf{\Sigma}$ whose first two eigenvectors $\boldsymbol{v}_j := (v_{j1}, \ldots, v_{jd})^T$ are specified to be sparse:

$$v_{1j} = \begin{cases} \frac{1}{\sqrt{10}} & 1 \le j \le 10 \\ 0 & \text{otherwise} \end{cases} \quad \text{and} \quad v_{2j} = \begin{cases} \frac{1}{\sqrt{10}} & 11 \le j \le 20 \\ 0 & \text{otherwise} \end{cases}.$$

Then we let $\mathbf{\Sigma}$ be $\mathbf{\Sigma} = 5\boldsymbol{v}_1\boldsymbol{v}_1^T + 2\boldsymbol{v}_2\boldsymbol{v}_2^T + \mathbf{I}_d$, where $\mathbf{I}_d \in \mathbb{R}^{d\times d}$ is the identity matrix. We have $\lambda_1(\mathbf{\Sigma}) = 6$, $\lambda_2(\mathbf{\Sigma}) = 3$, $\lambda_3(\mathbf{\Sigma}) = \ldots = \lambda_d(\mathbf{\Sigma}) = 1$. Using $\mathbf{\Sigma}$ as the covariance matrix, we generate $n$ data points from a Gaussian distribution or a multivariate-$t$ distribution with degrees of freedom 3. Here the dimension $d$ varies from 64 to 256 and the sample size $n$ varies from 10 to 500. Figure 1 plots the averaged angle distances $|\sin\angle(\widetilde{\boldsymbol{v}}_1, \boldsymbol{v}_1)|$ between the sparse ECA estimate $\widetilde{\boldsymbol{v}}_1$ and the true parameter $\boldsymbol{v}_1$, for dimensions $d = 64, 100, 256$, over 1,000 replications. In each setting, $s := \|\boldsymbol{v}_1\|_0$ is fixed to be a constant $s = 10$.

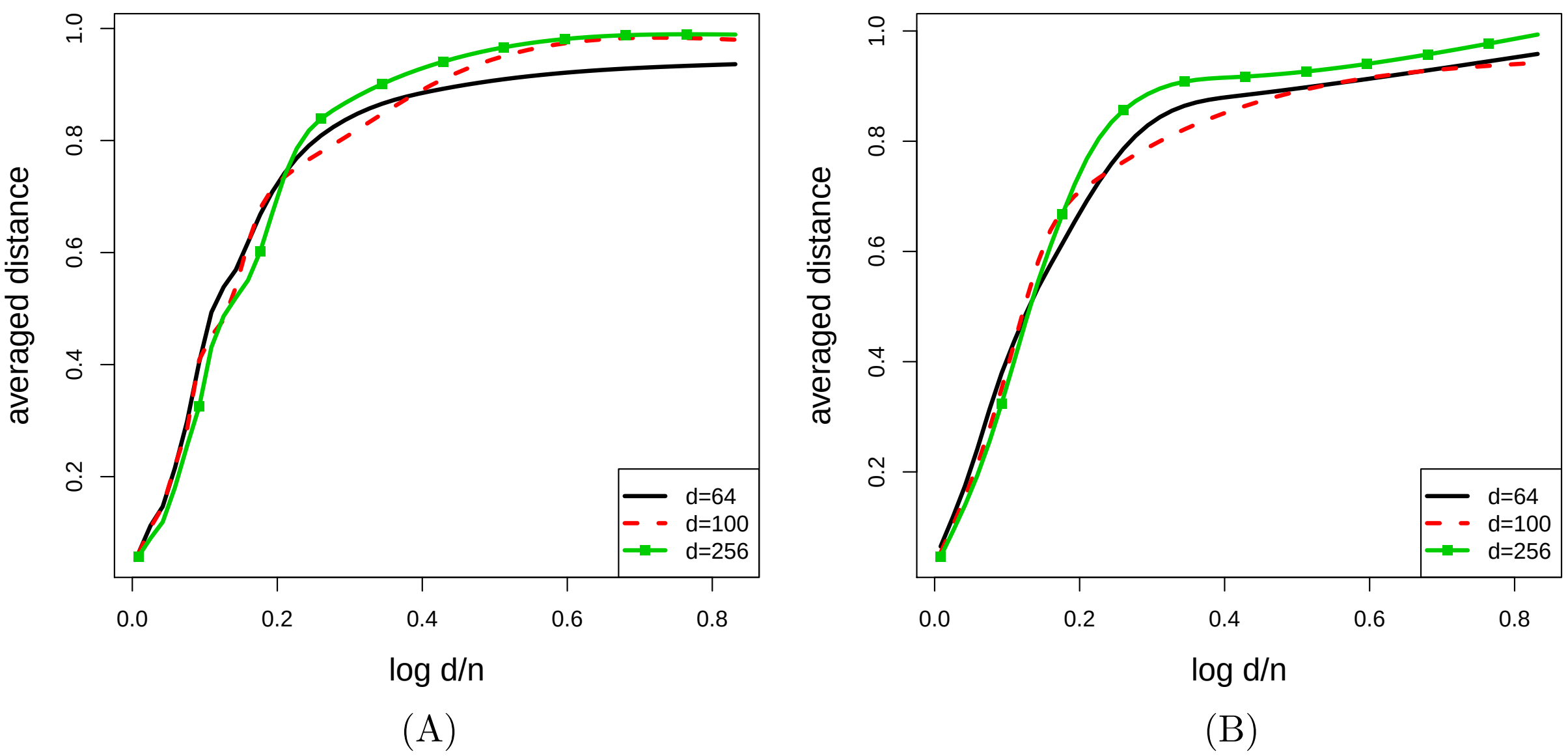

Figure 1: Simulation for two different distributions (normal and multivariate-$t$) with varying numbers of dimension $d$ and sample size $n$. Plots of averaged distances between the estimators and the true parameters are conducted over 1,000 replications. (A) Normal distribution; (B) Multivariate-$t$ distribution.

By examining the two curves in Figure 1 (A) and (B), the averaged distance between $\boldsymbol{v}_1$ and $\widetilde{\boldsymbol{v}}_1$ starts at almost zero (for sample size $n$ large enough), and then transits to almost one as the sample size decreases (in another word, $1/n$ increases simultaneously). Figure 1 shows that all curves almost overlapped with each other when the averaged distances are plotted against $\log d/n$. This phenomenon confirms the results in Theorem 5.4. Consequently, the ratio $n/\log d$ acts as an effective sample size in controlling the prediction accuracy of the eigenvectors.

In the supplementary materials, we further provide results when $s$ is set to be 5 and 20. There one will see the conclusion drawn here still holds.

#### 6.1.2 Estimating the Leading Eigenvector of the Covariance Matrix

We now focus on estimating the leading eigenvector of the covariance matrix $\boldsymbol{\Sigma}$. The first three rows in Table 2 list the simulation schemes of $(n, d)$ and $\boldsymbol{\Sigma}$. In detail, let $\omega_1 > \omega_2 > \omega_3 = \ldots = \omega_d$ be the eigenvalues and $\boldsymbol{v}_1, \ldots, \boldsymbol{v}_d$ be the eigenvectors of $\boldsymbol{\Sigma}$ with $\boldsymbol{v}_j := (v_{j1}, \ldots, v_{jd})^T$. The top $m$ leading eigenvectors $\boldsymbol{v}_1, \ldots, \boldsymbol{v}_m$ of $\boldsymbol{\Sigma}$ are specified to be sparse such that $s_j := \|\boldsymbol{v}_j\|_0$ is small and

$$v_{jk} = \begin{cases} 1/\sqrt{s_j}, & 1+\sum_{i=1}^{j-1} s_i \le k \le \sum_{i=1}^{j} s_i, \\ 0, & \text{otherwise.} \end{cases}$$

Table 2: Simulation schemes with different $n, d$ and $\mathbf{\Sigma}$. Here the eigenvalues of $\mathbf{\Sigma}$ are set to be $\omega_1 > \ldots > \omega_m > \omega_{m+1} = \ldots = \omega_d$ and the top $m$ leading eigenvectors $\boldsymbol{v}_1, \ldots, \boldsymbol{v}_m$ of $\mathbf{\Sigma}$ are specified to be sparse with $s_j := \|\boldsymbol{v}_j\|_0$ and $u_{jk} = 1/\sqrt{s_j}$ for $k \in [1 + \sum_{i=1}^{j-1} s_i, \sum_{i=1}^{j} s_i]$ and zero for all the others. $\mathbf{\Sigma}$ is generated as $\mathbf{\Sigma} = \sum_{j=1}^{m} (\omega_j - \omega_d) \boldsymbol{v}_j \boldsymbol{v}_j^T + \omega_d \mathbf{I}_d$. The column "Cardinalities" shows the cardinality of the support set of $\{\boldsymbol{v}_j\}$ in the form: "$s_1, s_2, \ldots, s_m, *, *, \ldots$". The column "Eigenvalues" shows the eigenvalues of $\mathbf{\Sigma}$ in the form: "$\omega_1, \omega_2, \ldots, \omega_m, \omega_d, \omega_d, \ldots$". In the first three schemes, $m$ is set to be 2; In the second three schemes, $m$ is set to be 4.

| Scheme | $n$ | $d$ | Cardinalities | Eigenvalues |
|---|---|---|---|---|
| Scheme 1 | 50 | 100 | $10, 10, *, *, \ldots$ | $6, 3, 1, 1, 1, 1, \ldots$ |
| Scheme 2 | 100 | 100 | $10, 10, *, * \ldots$ | $6, 3, 1, 1, 1, 1, \ldots$ |
| Scheme 3 | 100 | 200 | $10, 10, *, *, \ldots$ | $6, 3, 1, 1, 1, 1, \ldots$ |
| Scheme 4 | 50 | 100 | $10, 8, 6, 5, *, *, \ldots$ | $8, 4, 2, 1, 0.01, 0.01, \ldots$ |
| Scheme 5 | 100 | 100 | $10, 8, 6, 5, *, *, \ldots$ | $8, 4, 2, 1, 0.01, 0.01, \ldots$ |
| Scheme 6 | 100 | 200 | $10, 8, 6, 5, *, *, \ldots$ | $8, 4, 2, 1, 0.01, 0.01, \ldots$ |

Accordingly, $\mathbf{\Sigma}$ is generated as

$$\mathbf{\Sigma} = \sum_{j=1}^{m} (\omega_j - \omega_d) \boldsymbol{v}_j \boldsymbol{v}_j^T + \omega_d \mathbf{I}_d.$$

Table 2 shows the cardinalities $s_1, \ldots, s_m$ and eigenvalues $\omega_1, \ldots, \omega_m$ and $\omega_d$. In this section we set $m = 2$ (for the first three schemes) and $m = 4$ (for the later three schemes).

We consider the following four different elliptical distributions:

**(Normal)** $\boldsymbol{X} \sim EC_d(\mathbf{0}, \mathbf{\Sigma}, \xi_1 \cdot \sqrt{d/\mathbb{E}\xi_1^2})$ with $\xi_1 \stackrel{\mathsf{d}}{=} \chi_d$. Here $\chi_d$ is the chi-distribution with degrees of freedom $d$. For $Y_1, \ldots, Y_d \stackrel{i.i.d.}{\sim} N(0, 1)$,

$$\sqrt{Y_1^2 + \ldots + Y_d^2} \stackrel{\mathsf{d}}{=} \chi_d.$$

In this setting, $\boldsymbol{X}$ follows a Gaussian distribution (Fang et al., 1990).

**(Multivariate-$t$)** $\boldsymbol{X} \sim EC_d(\mathbf{0}, \mathbf{\Sigma}, \xi_2 \cdot \sqrt{d/\mathbb{E}\xi_2^2})$ with $\xi_2 \stackrel{\mathsf{d}}{=} \sqrt{\kappa}\xi_1^*/\xi_2^*$. Here $\xi_1^* \stackrel{\mathsf{d}}{=} \chi_d$ and $\xi_2^* \stackrel{\mathsf{d}}{=} \chi_\kappa$ with $\kappa \in \mathbb{Z}^+$. In this setting, $\boldsymbol{X}$ follows a multivariate-$t$ distribution with degrees of freedom $\kappa$ (Fang et al., 1990). Here we consider $\kappa = 3$.

**(EC1)** $\boldsymbol{X} \sim EC_d(\mathbf{0}, \mathbf{\Sigma}, \xi_3)$ with $\xi_3 \sim F(d, 1)$, i.e., $\xi_3$ follows an $F$-distribution with degrees of freedom $d$ and 1. Here $\xi_3$ has no finite mean. But ECA could still estimate the eigenvectors of the scatter matrix and is thus robust.

**(EC2)** $\boldsymbol{X} \sim EC_d(\mathbf{0}, \mathbf{\Sigma}, \xi_4 \cdot \sqrt{d/\mathbb{E}\xi_4^2})$ with $\xi_4 \sim \text{Exp}(1)$, i.e., $\xi_4$ follows an exponential distribution with the rate parameter 1.

We generate $n$ data points according to the schemes 1 to 3 and the four distributions discussed above 1,000 times each. To show the estimation accuracy, Figure 2 plots the averaged distances between the estimate $\widehat{\boldsymbol{v}}_1$ and $\boldsymbol{v}_1$, defined as $|\sin \angle(\widehat{\boldsymbol{v}}_1, \boldsymbol{v}_1)|$, against the number of estimated nonzero entries (defined as $\|\widehat{\boldsymbol{v}}_1\|_0$), for three different methods: $\mathsf{TP}$,$\mathsf{TCA}$, and $\mathsf{ECA}$.

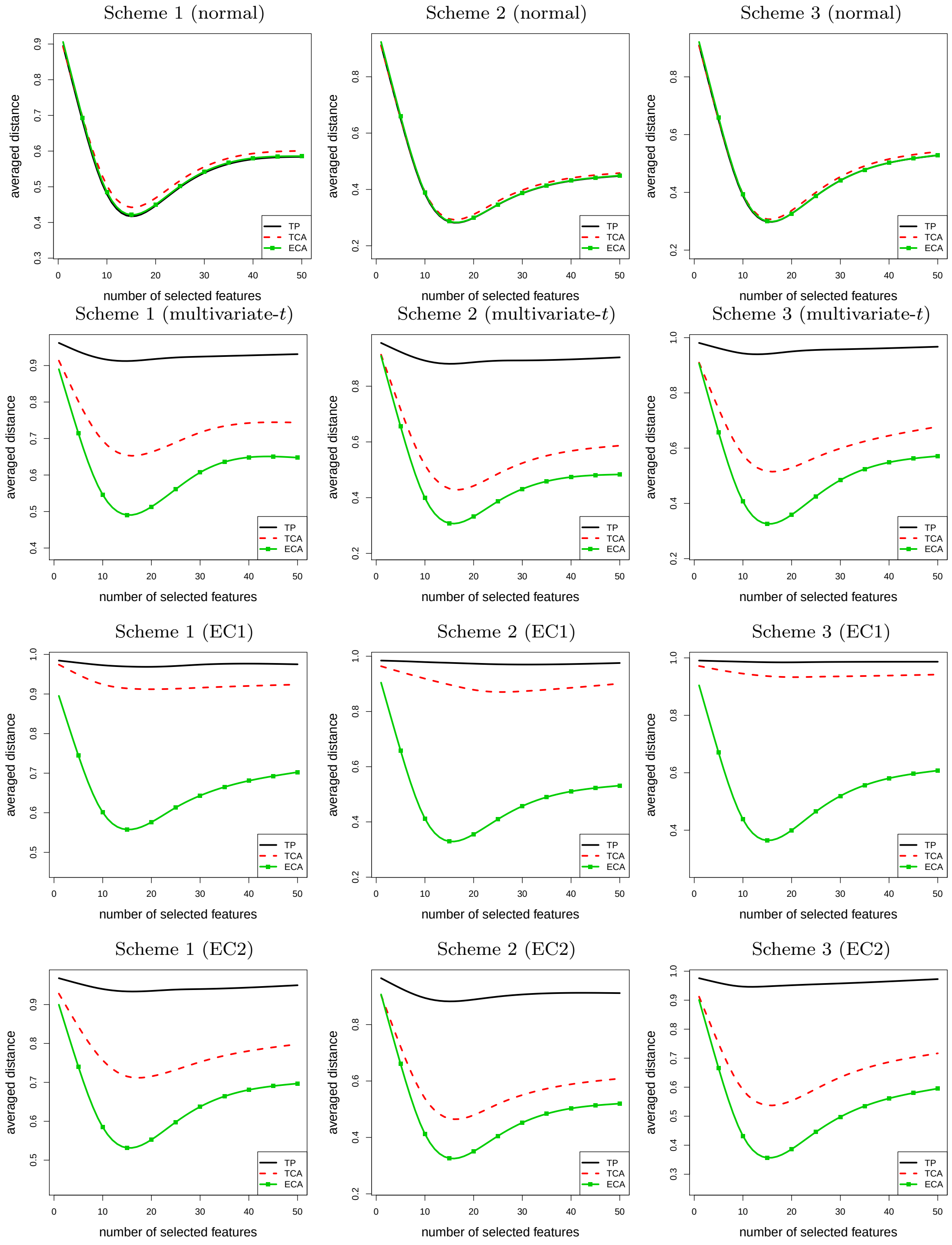

Figure 2: Curves of averaged distances between the estimates and true parameters for different schemes and distributions (normal, multivariate-$t$, EC1, and EC2, from top to bottom) using the FTPM algorithm. Here we are interested in estimating the leading eigenvector. The horizontal-axis represents the cardinalities of the estimates' support sets and the vertical-axis represents the averaged distances.

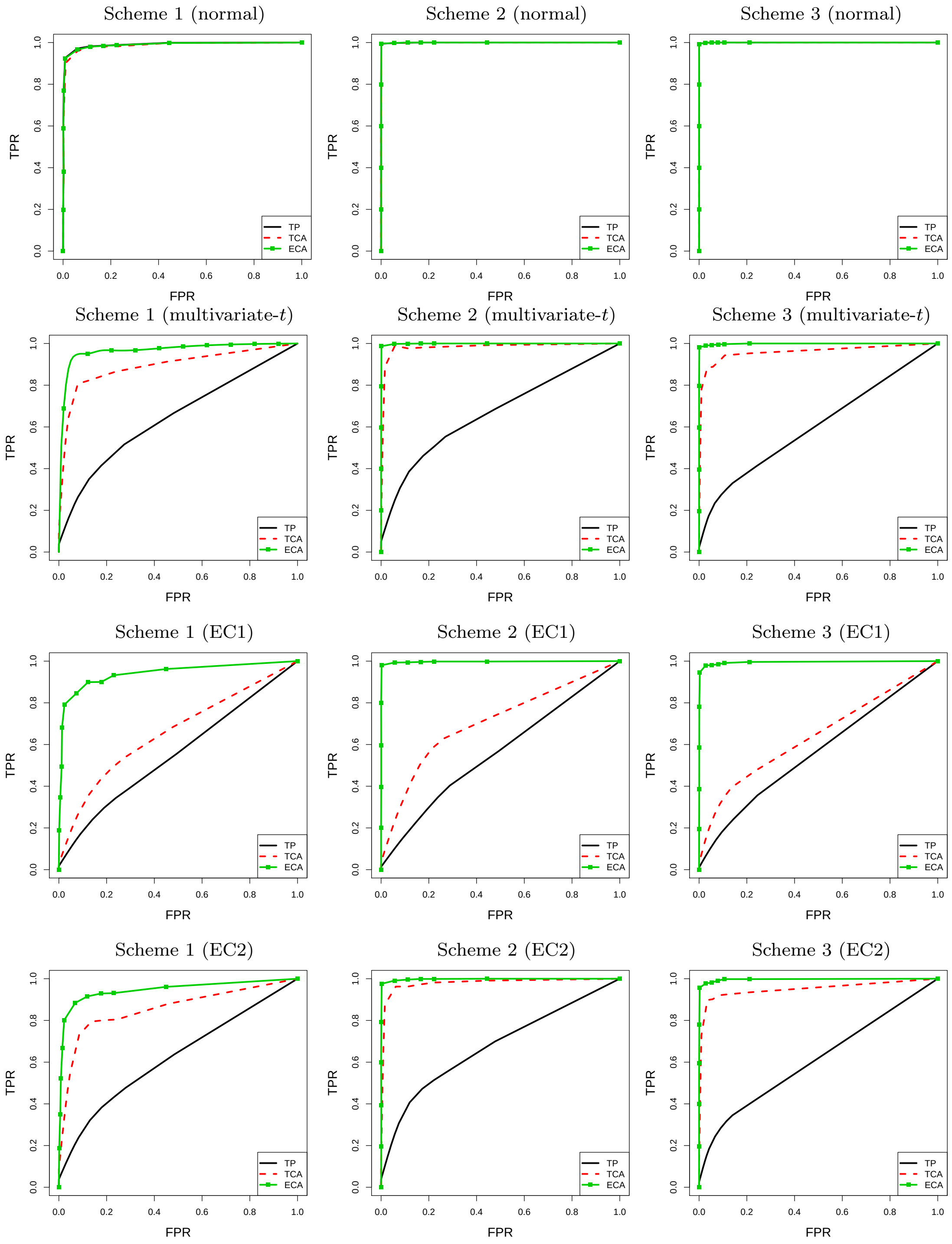

Figure 3: ROC curves for different methods in schemes 1 to 3 and different distributions (normal, multivariate-$t$, EC1, and EC2, from top to bottom) using the FTPM algorithm. Here we are interested in estimating the sparsity pattern of the leading eigenvector.

Table 3: Testing for normality of the ABIDE data. This table illustrates the number of voxels (out of a total number 116) rejecting the null hypothesis of normality at the significance level of 0.05 with or without Bonferroni's adjustment.

| Critical value | **Kolmogorov-Smirnov** | **Shapiro-Wilk** | **Lilliefors** |
|---|---|---|---|
| 0.05 | 88 | 115 | 115 |
| 0.05/116 | 61 | 113 | 92 |

To show the feature selection results for estimating the support set of the leading eigenvector $\boldsymbol{v}_1$, Figure 3 plots the false positive rates against the true positive rates for the three different estimators under different schemes of $(n, d)$, $\boldsymbol{\Sigma}$, and different distributions.

Figure 2 shows that when the data are non-Gaussian but follow an elliptical distribution, ECA consistently outperforms TCA and TP in estimation accuracy. Moreover, when the data are indeed normal, there is no obvious difference between ECA and TP, indicating that ECA is a safe alternative to sparse PCA within the elliptical family. Furthermore, Figure 3 verifies that, in term of feature selection, the same conclusion can be drawn.

In the supplementary materials, we also provide results when the data are Cauchy distributed and the same conclusion holds.

#### 6.1.3 Estimating the Top $m$ Leading Eigenvectors of the Covariance Matrix

Next, we focus on estimating the top $m$ leading eigenvectors of the covariance matrix $\boldsymbol{\Sigma}$. We generate $\boldsymbol{\Sigma}$ in a similar way as in Section 6.1.2. We adopt the schemes 4 to 6 in Table 2 and the four distributions discussed in Section 6.1.2. We consider the case $m = 4$. We use the iterative deflation method and exploit the FTPM algorithm in each step to estimate the eigenvectors $\boldsymbol{v}_1, \ldots, \boldsymbol{v}_4$. The tuning parameter remains the same in each iterative deflation step.

Parallel to the last section, Figure 4 plots the distances between the estimates $\widehat{\boldsymbol{v}}_1, \ldots, \widehat{\boldsymbol{v}}_4$ and the true parameters $\boldsymbol{v}_1, \ldots, \boldsymbol{v}_4$ against the numbers of estimated nonzero entries. Here the distance is defined as $\sum_{j=1}^4 |\sin\angle(\boldsymbol{v}_j, \widehat{\boldsymbol{v}}_j)|$ and the number is defined as $\sum_{j=1}^4 \|\widehat{\boldsymbol{v}}_j\|_0$. We see that the averaged distance starts at 4, decreases first and then increases with the number of estimated nonzero entries. The minimum is achieved when the number of nonzero entries is 40. The same conclusion drawn in the last section holds here, indicating that ECA is a safe alternative to sparse PCA when the data are elliptically distributed.

### 6.2 Brain Imaging Data Study

In this section we apply ECA and the other two methods to a brain imaging data obtained from the Autism Brain Imaging Data Exchange (ABIDE) project (http://fcon_1000.projects.nitrc.org/indi/abide/). The ABIDE project shares over 1,000 functional and structural scans for individuals with and without autism. This dataset includes 1,043 subjects, of which 544 are the controls and the rest are diagnosed with autism. Each subject is scanned at multiple time points, ranging from 72 to 290. The data were pre-processed for correcting motion and eliminating noise. We refer to Di Martino et al. (2014) and Kang (2013) for more detail on data preprocessing procedures.

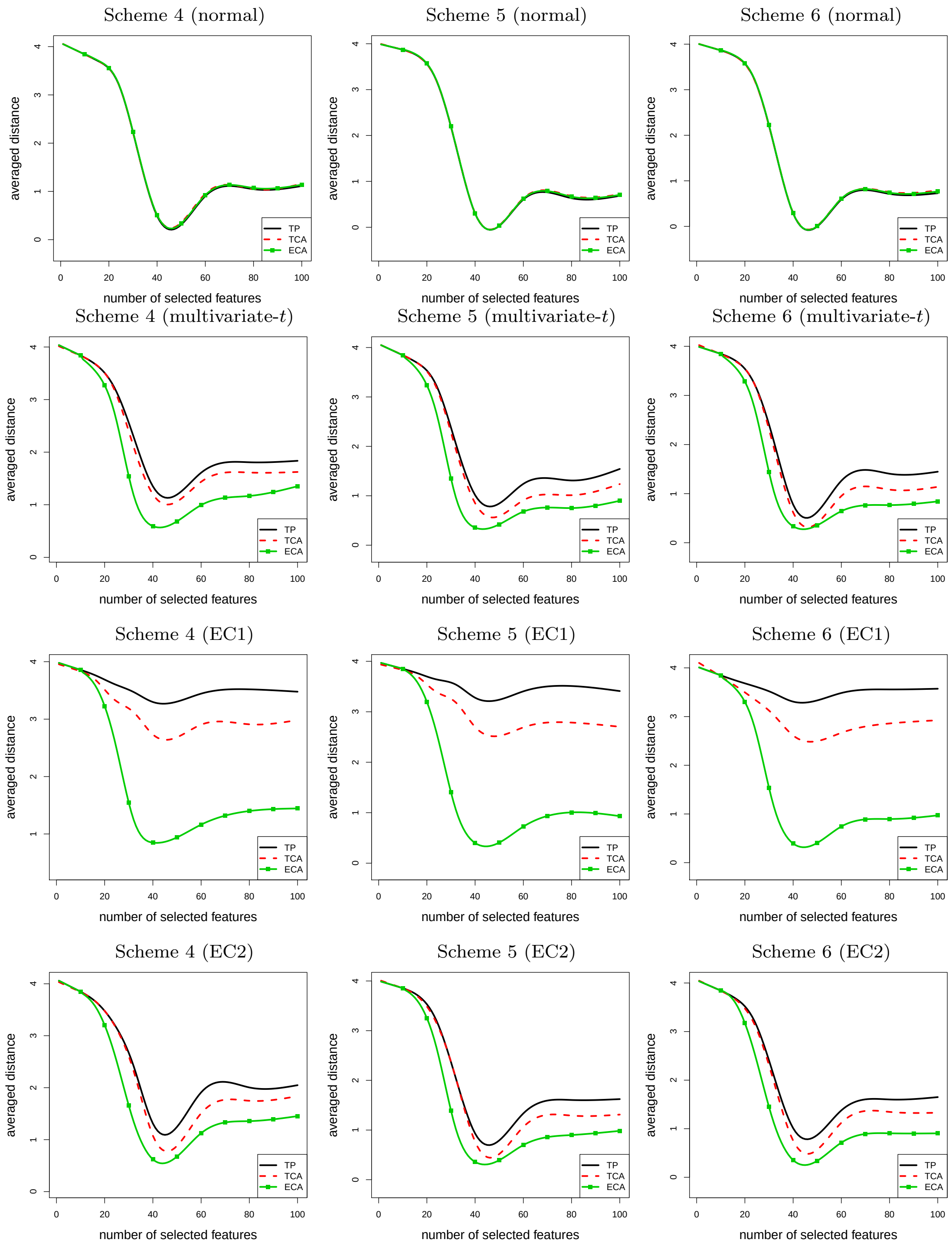

Figure 4: Curves of averaged distances between the estimates and true parameters for different methods in schemes 4 to 6 and different distributions (normal, multivariate-$t$, EC1, and EC 2, from top to bottom) using the FTPM algorithm. Here we are interested in estimating the top 4 leading eigenvectors. The horizontal-axis represents the cardinalities of the estimates' support sets and the vertical-axis represents the averaged distances.

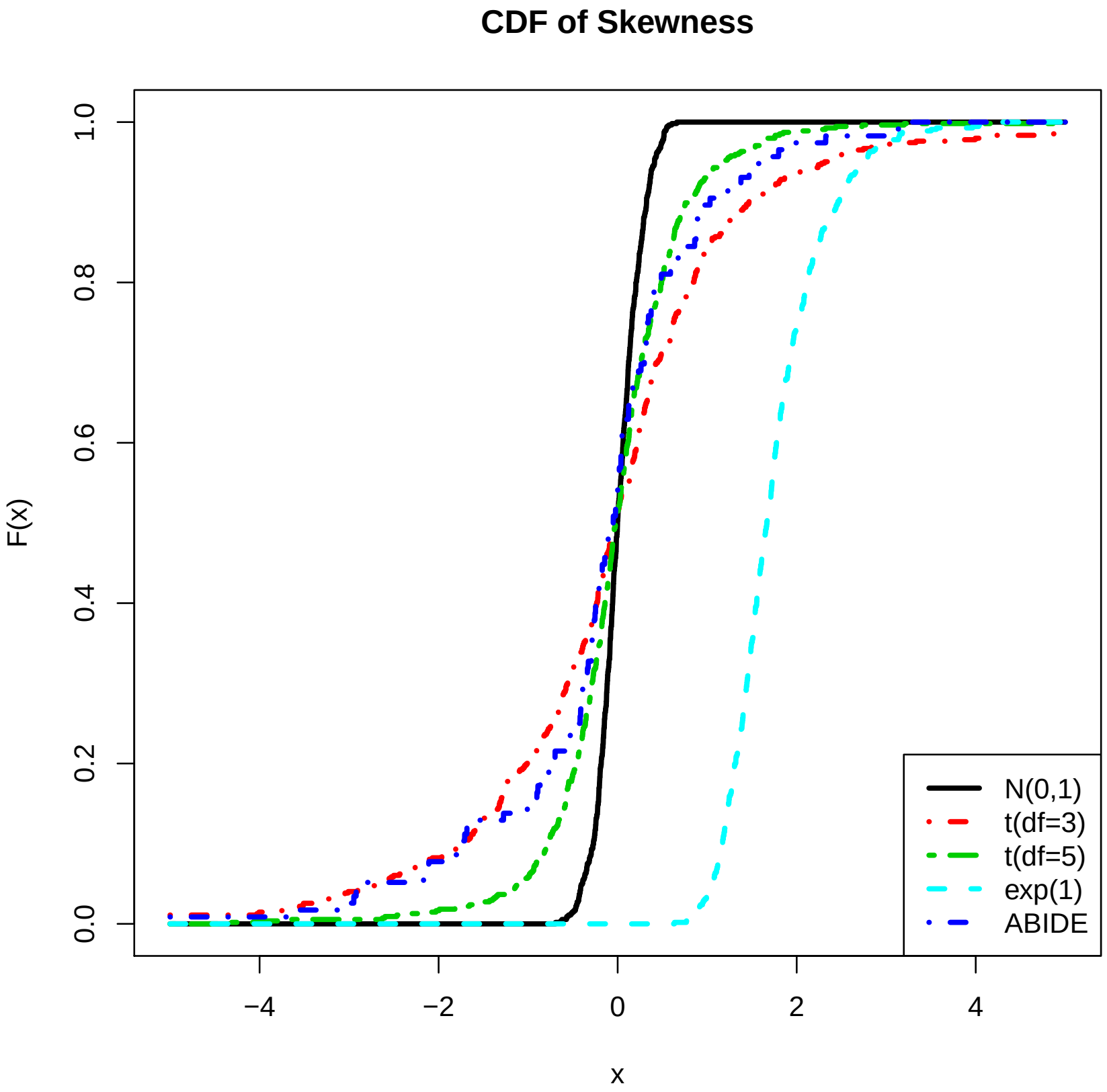

Figure 5: Illustration of the symmetric and heavy-tailed properties of the brain imaging data. The estimated cumulative distribution functions (CDF) of the marginal skewness based on the ABIDE data and four simulated distributions are plotted against each other.

Based on the 3D scans, we extract 116 regions that are of interest from the AAL atlas (Tzourio-Mazoyer et al., 2002) and broadly cover the brain. This gives us 1,043 matrices, each with 116 columns and number of rows from 72 to 290. We then followed the idea in Eloyan et al. (2012) and Han et al. (2013) to compress the information of each subject by taking the median of each column for each matrix. In this study, we are interested in studying the control group. This gives us a $544 \times 116$ matrix.

First, we explore the obtained dataset to unveil several characteristics. In general, we find that the observed data are non-Gaussian and marginally symmetric. We first illustrate the non-Gaussian issue. Table 3 provides the results of marginal normality tests. Here we conduct the three marginal normality tests at the significant level of 0.05. It is clear that at most 28 out of 116 voxels would pass any of three normality test. Even with Bonferroni correction, over half the voxels fail to pass any normality tests. This indicates that the imaging data are not Gaussian distributed.

We then show that the data are marginally symmetric. For this, we first calculate the marginal skewness of each column in the data matrix. We then compare the empirical distribution function based on the marginal skewness values of the data matrix with that based on the simulated data from the standard Gaussian ($N(0,1)$), $t$ distribution with degree freedom 3 ($t(\text{df}=3)$), $t$ distribution with degree freedom 5 ($t(\text{df}=5)$), and the exponential distribution with the rate parameter 1 ($\exp(1)$). Here the first three distributions are symmetric and the exponential distribution is skewed to the right. Figure 5 plots the five estimated distribution functions. We see that the distribution function for the marginal skewness of the imaging data is very close to that of the $t(\text{df}=3)$ distribution . This indicates that the data are marginally symmetric. Moreover, the distribution function based on the imaging data is far away from that based on the Gaussian distribution, indicating that the data can be heavy-tailed.

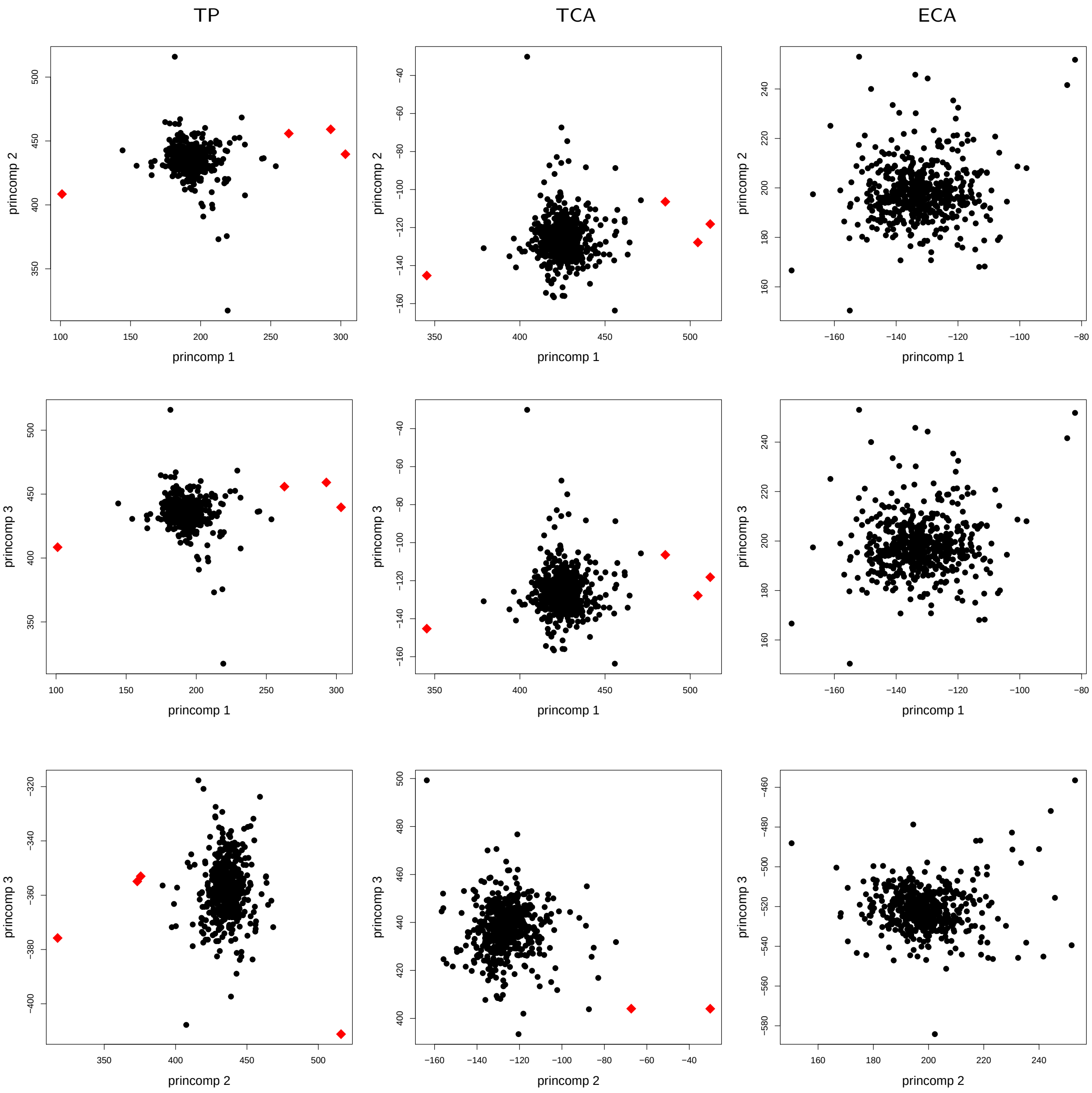

Figure 6: Plots of principal components 1 against 2, 1 against 3, 2 against 3 from top to bottom. The methods used are TP, TCA and ECA. Here red dots represent the points with strong leverage influence.

The above data exploration reveals that the ABIDE data are non-Gaussian, symmetric, and heavy-tailed, which makes the elliptical distribution very appealing to model the data. We then apply TP, TCA and ECA to this dataset. We extract the top three eigenvectors and set the tuning parameter of the truncated power method to be 40. We project each pair of principal components of the ABIDE data onto 2D plots, shown in Figure 6. Here the red dots represent the possible outliers that have strong leverage influence. The leverage strength is defined as the diagonal values of the hat matrix in the linear model by regressing the first principal component on the second one (Neter et al., 1996). High leverage strength means that including these points will severely affect the linear regression estimates applied to principal components of the data. A data point is said to have strong leverage influence if its leverage strength is higher than a chosen threshold value. Here we choose the threshold value to be $0.05(\approx 27/544)$.

It can be observed that there are points with strong leverage influence for both statistics learnt by TP and TCA, while none for ECA. This implies that ECA has the potential to deliver better results for inference based on the estimated principal components.

# ECA: High Dimensional Elliptical Component Analysis in non-Gaussian Distributions (Supplementary Appendix)

Fang Han* and Han Liu†

The supplementary materials provide more simulation results, as well as all technical proofs.

## A More Simulation Results

This section provides more simulation results.

### A.1 More Results in Section 6.1.1

Following the results in Section 6.1.1, we evaluate ECA's dependence on sample size and dimension with the sparsity values $s = 5$ and 20. Here all the nonzero entries in $\boldsymbol{v}_1$ and $\boldsymbol{v}_2$ are set to be equal to $1/\sqrt{s}$. All the other parameters remain same as in Section 6.1.1. The corresponding results are put in Figures 1, and the conclusion drawn in Section 6.1.1 still holds here.

### A.2 More Results in Section 6.1.2

Following the results in Section 6.1.2, we further evaluate the performance of ECA and its competitors when the data are Cauchy distributed. In particular, we consider the following setting:

**(Cauchy)** $\boldsymbol{X} \sim EC_d(\mathbf{0}, \boldsymbol{\Sigma}, \xi_2\sqrt{d})$ with $\xi_2 \overset{\mathsf{d}}{=} \xi_1^*/\xi_2^*$. Here $\xi_1^* \overset{\mathsf{d}}{=} \chi_d$ and $\xi_2^* \overset{\mathsf{d}}{=} \chi_1$.

All the other parameters remain same as in Schemes 1, 2, 3 in Section 6.1.2. Figures 2 and 3 illustrate the estimation and model selection efficiency of the competing methods. It could be observed that ECA's performance remains best. Actually, its advantage over TP and TCA is more significant than that under the multivariate-$t$ with the degree of freedom 3. This is as expected since the model is even more heavy-tailed now. The results also, similar to the case EC1, empirically verify that ECA works well even if the covariance matrix does not exist.

*Department of Statistics, University of Washington, Seattle, WA 98195, USA; e-mail: fanghan@uw.edu

†Department of Operations Research and Financial Engineering, Princeton University, Princeton, NJ 08544, USA; e-mail: hanliu@princeton.edu

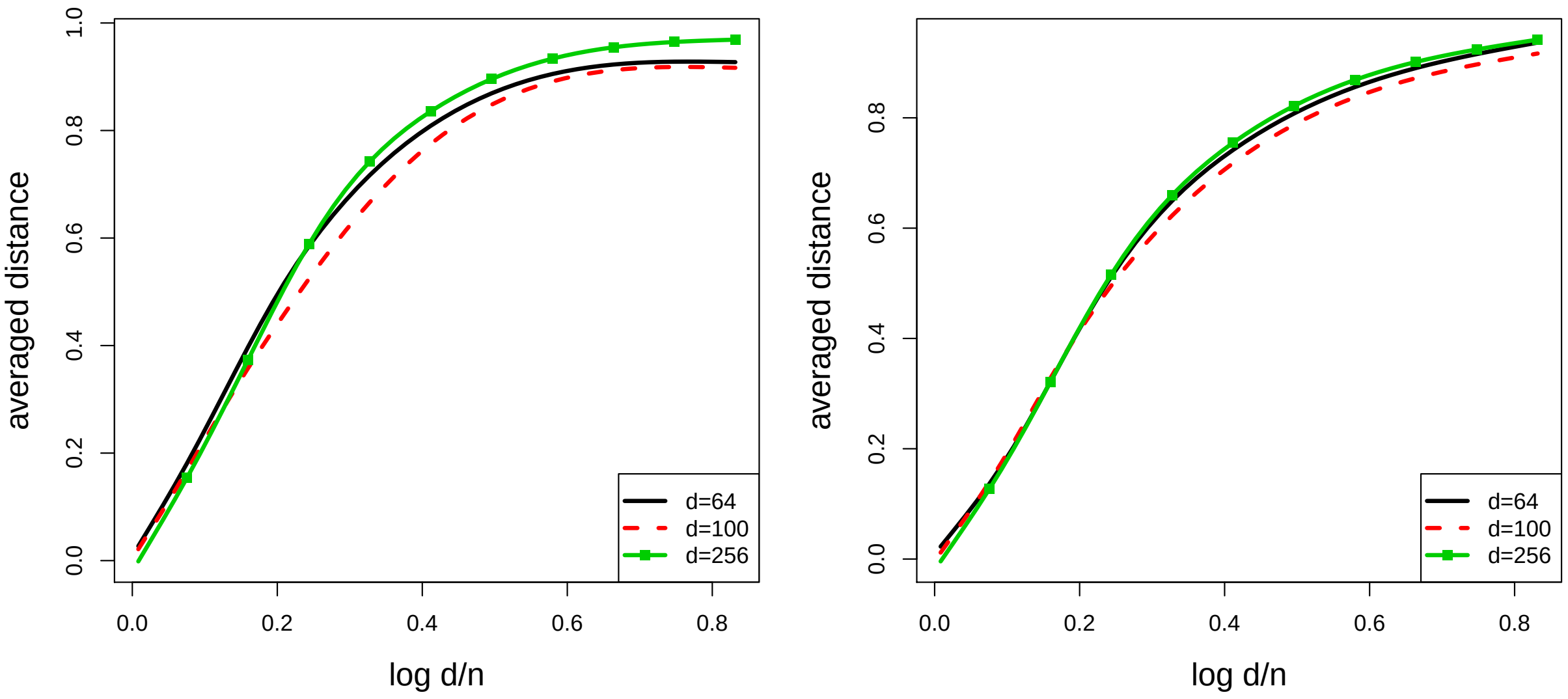

(A) (B)

Figure 1: Simulation for multivariate-$t$ with varying numbers of dimension $d$ and sample size $n$. Plots of averaged distances between the estimators and the true parameters are conducted over 1,000 replications. (A) Multivariate-$t$ distribution with $s = 5$; (B) Multivariate-$t$ distribution with $s = 20$.

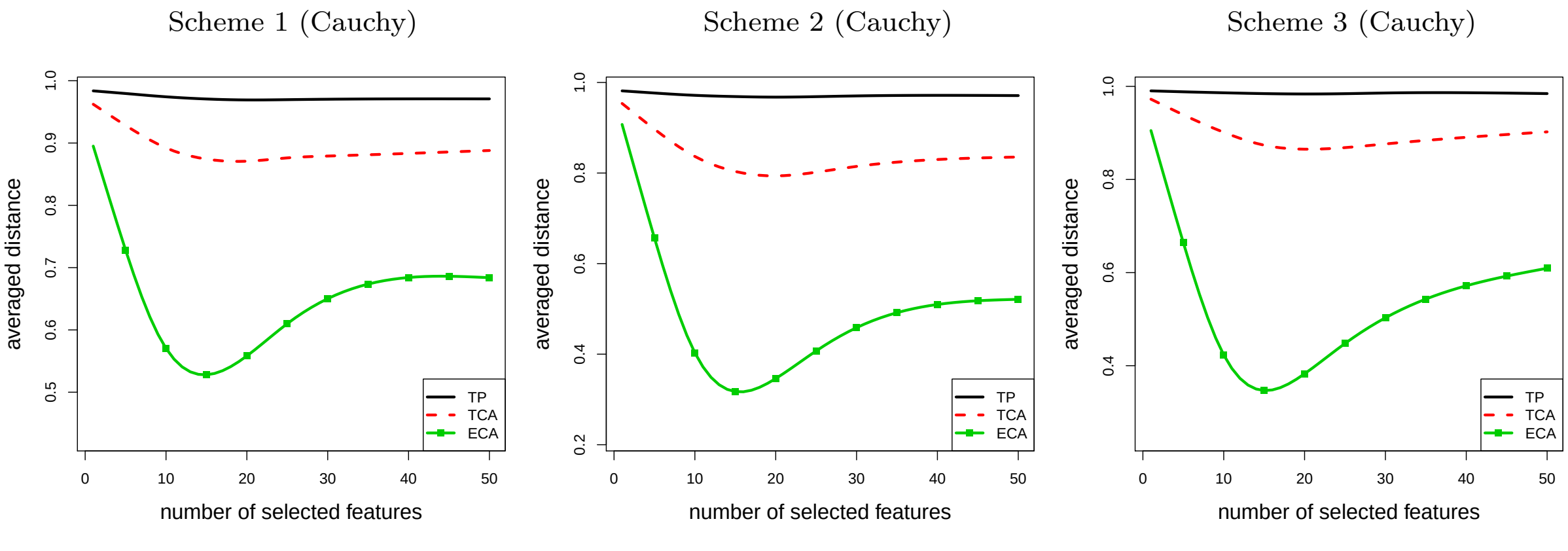

Figure 2: Curves of averaged distances between the estimates and true parameters for different schemes and Cauchy distribution using the FTPM algorithm. Here we are interested in estimating the leading eigenvector. The horizontal-axis represents the cardinalities of the estimates' support sets and the vertical-axis represents the averaged distances.

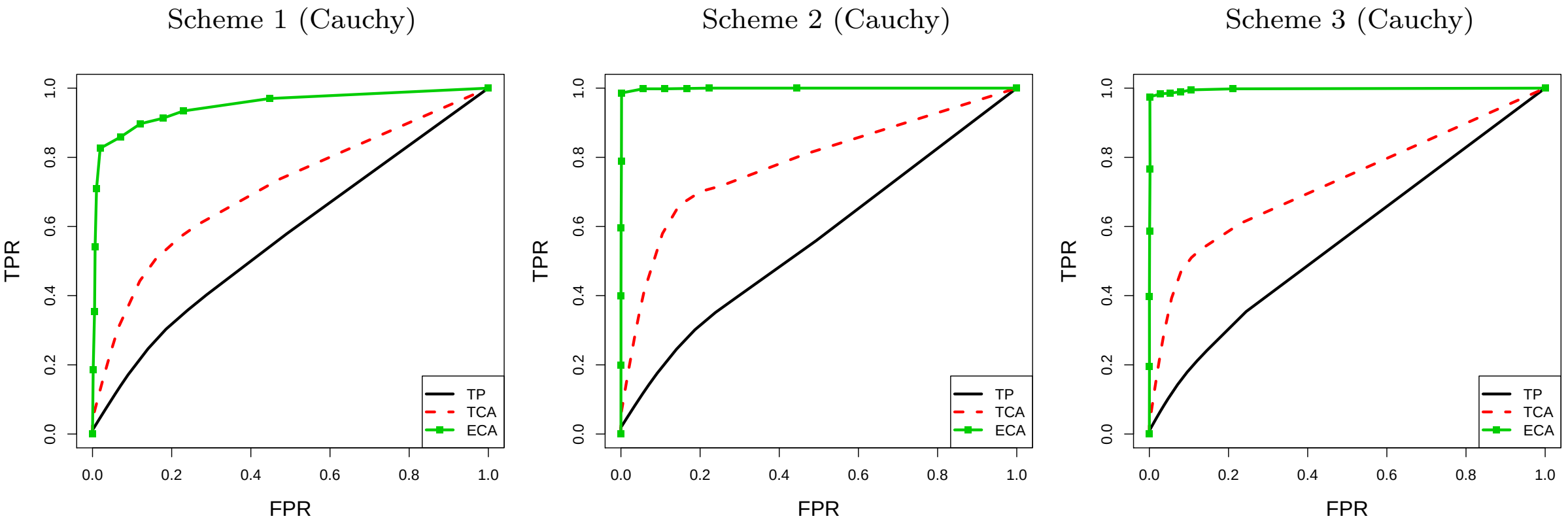

Figure 3: ROC curves for different methods in schemes 1 to 3 and Cauchy distributions using the FTPM algorithm. Here we ar interested in estimating the sparsity pattern of the leading eigenvector.

# B Proofs

In this section we provide the proofs of results shown in Sections 2, 3, 4, and 5.

## B.1 Proofs of Results in Section 2

This section proves Proposition 2.1. The proof summarizes the results in Marden (1999) and Croux et al. (2002), and is provided only for completeness. In particular, we do not claim any original contribution.

**Lemma B.1.** Let $\boldsymbol{X} \sim EC_d(\boldsymbol{\mu}, \boldsymbol{\Sigma}, \xi)$ be a continuous random vector. We have

$$\mathbf{K} = \mathbb{E}\left(\frac{(\boldsymbol{X}-\widetilde{\boldsymbol{X}})(\boldsymbol{X}-\widetilde{\boldsymbol{X}})^T}{\|\boldsymbol{X}-\widetilde{\boldsymbol{X}}\|_2^2}\right) = \mathbb{E}\left(\frac{(\boldsymbol{X}-\boldsymbol{\mu})(\boldsymbol{X}-\boldsymbol{\mu})^T}{\|\boldsymbol{X}-\boldsymbol{\mu}\|_2^2}\right). \tag{B.1}$$

*Proof.* By the equivalent definition of the elliptical distribution, there exists a characteristic function $\psi$ uniquely determined by $\xi$ such that $\boldsymbol{X} \sim EC_d(\boldsymbol{\mu}, \boldsymbol{\Sigma}, \psi)$ and $\widetilde{\boldsymbol{X}} \sim EC_d(\boldsymbol{\mu}, \boldsymbol{\Sigma}, \psi)$. Let $i := \sqrt{-1}$. Since $\boldsymbol{X}$ and $\widetilde{\boldsymbol{X}}$ are independent, we have $\mathbb{E}\exp(i\boldsymbol{t}^T(\boldsymbol{X}-\widetilde{\boldsymbol{X}})) = \mathbb{E}\exp(i\boldsymbol{t}^T\boldsymbol{X})\mathbb{E}\exp(-i\boldsymbol{t}^T\widetilde{\boldsymbol{X}}) = \psi^2(\boldsymbol{t}^T\boldsymbol{\Sigma}\boldsymbol{t})$, implying that $\boldsymbol{X}-\widetilde{\boldsymbol{X}} \sim EC_d(\boldsymbol{0}, \boldsymbol{\Sigma}, \boldsymbol{\psi}^2)$. Again, by the equivalent definition of the elliptical distribution, there exists a nonnegative random variable $\xi'$ uniquely determined by $\psi^2$, such that $\boldsymbol{X}-\widetilde{\boldsymbol{X}} \sim EC_d(\boldsymbol{0}, \boldsymbol{\Sigma}, \xi')$. Because $\boldsymbol{X}$ is continuous, we have $\mathbb{P}(\xi' = 0) = 0$. Therefore,

$$\begin{aligned}\mathbf{K} &= \mathbb{E}\left(\frac{(\boldsymbol{X}-\widetilde{\boldsymbol{X}})(\boldsymbol{X}-\widetilde{\boldsymbol{X}})^T}{\|\boldsymbol{X}-\widetilde{\boldsymbol{X}}\|_2^2}\right) = \mathbb{E}\left(\frac{(\xi'\mathbf{A}\boldsymbol{U})(\xi'\mathbf{A}\boldsymbol{U})^T}{\|\xi'\mathbf{A}\boldsymbol{U}\|_2^2}\right) \\ &= \mathbb{E}\left(\frac{(\mathbf{A}\boldsymbol{U})(\mathbf{A}\boldsymbol{U})^T}{\|\mathbf{A}\boldsymbol{U}\|_2^2}\right) = \mathbb{E}\left(\frac{(\xi\mathbf{A}\boldsymbol{U})(\xi\mathbf{A}\boldsymbol{U})^T}{\|\xi\mathbf{A}\boldsymbol{U}\|_2^2}\right) = \mathbb{E}\left(\frac{(\boldsymbol{X}-\boldsymbol{\mu})(\boldsymbol{X}-\boldsymbol{\mu})^T}{\|\boldsymbol{X}-\boldsymbol{\mu}\|_2^2}\right).\end{aligned}$$

This completes the proof. □

*Proof of Proposition 2.1.* Using Lemma B.1, it is equivalent to consider $\mathbf{K} = \mathbb{E}\left(\frac{(\boldsymbol{X}-\boldsymbol{\mu})(\boldsymbol{X}-\boldsymbol{\mu})^T}{\|\boldsymbol{X}-\boldsymbol{\mu}\|_2^2}\right)$. Letting $\boldsymbol{\Omega} := [\boldsymbol{u}_1(\boldsymbol{\Sigma}), \ldots, \boldsymbol{u}_d(\boldsymbol{\Sigma})]$, $\boldsymbol{u}_{q+1}(\boldsymbol{\Sigma})$ until $\boldsymbol{u}_d(\boldsymbol{\Sigma})$ chosen to be orthogonal to $\boldsymbol{u}_1(\boldsymbol{\Sigma}), \ldots, \boldsymbol{u}_q(\boldsymbol{\Sigma})$

(which are also specified to be orthogonal to each other), we have

$$\|\boldsymbol{X}-\boldsymbol{\mu}\|_2 = \|\boldsymbol{\Omega}^T(\boldsymbol{X}-\boldsymbol{\mu})\|_2.$$

This implies that

$$\boldsymbol{\Omega}^T\frac{\boldsymbol{X}-\boldsymbol{\mu}}{\|\boldsymbol{X}-\boldsymbol{\mu}\|_2} = \frac{\boldsymbol{\Omega}^T(\boldsymbol{X}-\boldsymbol{\mu})}{\|\boldsymbol{\Omega}^T(\boldsymbol{X}-\boldsymbol{\mu})\|_2} = \frac{\boldsymbol{Z}}{\|\boldsymbol{Z}\|_2},$$

where using the stochastic representation of $\boldsymbol{X}$ in Equation (2.1), we have $\boldsymbol{Z} = \boldsymbol{\Omega}^T\mathbf{A}\boldsymbol{U} = \mathbf{D}\boldsymbol{U}$ with $\mathbf{D} = (\mathrm{diag}(\sqrt{\lambda_1(\boldsymbol{\Sigma})},\ldots,\sqrt{\lambda_q(\boldsymbol{\Sigma})}),\mathbf{0})^T \in \mathbb{R}^{d\times q}$. Therefore,

$$\mathbf{K} = \mathbb{E}\frac{(\boldsymbol{X}-\boldsymbol{\mu})(\boldsymbol{X}-\boldsymbol{\mu})^T}{\|\boldsymbol{X}-\boldsymbol{\mu}\|_2^2} = \boldsymbol{\Omega}\cdot\left[\mathbb{E}\left(\frac{\boldsymbol{Z}\boldsymbol{Z}^T}{\|\boldsymbol{Z}\|_2^2}\right)\right]\cdot\boldsymbol{\Omega}^T.$$

Secondly, we prove that $\mathbb{E}\left(\frac{\boldsymbol{Z}\boldsymbol{Z}^T}{\|\boldsymbol{Z}\|_2^2}\right)$ is a diagonal matrix. This is because, for any matrix $\mathbf{P} = \mathrm{diag}(\boldsymbol{v})$, where $\boldsymbol{v} = (v_1,\ldots,v_d)^T$ satisfies that $v_j = 1$ or $-1$ for $j = 1,\ldots,d$, we have

$$\mathbf{P}\frac{\boldsymbol{Z}}{\|\boldsymbol{Z}\|_2} = \frac{\mathbf{P}\boldsymbol{Z}}{\|\mathbf{P}\boldsymbol{Z}\|_2} \stackrel{d}{=} \frac{\boldsymbol{Z}}{\|\boldsymbol{Z}\|_2} \quad\Rightarrow\quad \mathbb{E}\left(\frac{\boldsymbol{Z}\boldsymbol{Z}^T}{\|\boldsymbol{Z}\|_2^2}\right) = \mathbf{P}\left[\mathbb{E}\left(\frac{\boldsymbol{Z}\boldsymbol{Z}^T}{\|\boldsymbol{Z}\|_2^2}\right)\right]\mathbf{P}.$$

It holds if and only if $\mathbb{E}\left(\frac{\boldsymbol{Z}\boldsymbol{Z}^T}{\|\boldsymbol{Z}\|_2^2}\right)$ is a diagonal matrix.

To finish the proof, we need to show that the diagonals of $\mathbb{E}\left(\frac{\boldsymbol{Z}\boldsymbol{Z}^T}{\|\boldsymbol{Z}\|_2^2}\right)$ are decreasing. Reminding that $\boldsymbol{Z} = \mathbf{D}\boldsymbol{U}$, we have that

$$\mathbb{E}\left(\frac{\boldsymbol{Z}\boldsymbol{Z}^T}{\|\boldsymbol{Z}\|_2^2}\right) = \mathbb{E}\left(\frac{\mathbf{D}\boldsymbol{U}\boldsymbol{U}^T\mathbf{D}}{\boldsymbol{U}^T\mathbf{D}^2\boldsymbol{U}}\right).$$

Letting $\boldsymbol{U} := (U_1,\ldots,U_q)^T$, by algebra, for $j = 1,\ldots,q$,

$$\left[\mathbb{E}\left(\frac{\boldsymbol{Z}\boldsymbol{Z}^T}{\|\boldsymbol{Z}\|_2^2}\right)\right]_{jj} = \mathbb{E}\left(\frac{\lambda_j(\boldsymbol{\Sigma})U_j^2}{\lambda_1(\boldsymbol{\Sigma})U_1^2+\ldots+\lambda_q(\boldsymbol{\Sigma})U_q^2}\right).$$

Actually, we have for any $k < j$,

$$\frac{\lambda_k(\mathbf{K})}{\lambda_j(\mathbf{K})} = \frac{\mathbb{E}\frac{\lambda_k(\boldsymbol{\Sigma})U_k^2}{\lambda_j(\boldsymbol{\Sigma})U_j^2+\lambda_k(\boldsymbol{\Sigma})U_k^2+E}}{\mathbb{E}\frac{\lambda_j(\boldsymbol{\Sigma})U_j^2}{\lambda_j(\boldsymbol{\Sigma})U_j^2+\lambda_k(\boldsymbol{\Sigma})U_k^2+E}} < \frac{\mathbb{E}\frac{\lambda_k(\boldsymbol{\Sigma})U_k^2}{\lambda_k(\boldsymbol{\Sigma})U_j^2+\lambda_k(\boldsymbol{\Sigma})U_k^2+E}}{\mathbb{E}\frac{\lambda_j(\boldsymbol{\Sigma})U_j^2}{\lambda_j(\boldsymbol{\Sigma})U_j^2+\lambda_j(\boldsymbol{\Sigma})U_k^2+E}} = \frac{\mathbb{E}\frac{U_k^2}{U_j^2+U_k^2+E/\lambda_k(\boldsymbol{\Sigma})}}{\mathbb{E}\frac{U_k^2}{U_j^2+U_k^2+E/\lambda_j(\boldsymbol{\Sigma})}} < 1,$$

where we let $E := \sum_{i\notin\{j,k\}}\lambda_i(\boldsymbol{\Sigma})U_i^2$. This completes the proof. □

## B.2 Proofs of Results in Section 3

In this section we provide the proofs of Theorems 3.1 and 3.5. To prove Theorem 3.1, we exploit the U-statistics version of the matrix Bernstein's inequality (Tropp, 2012), which is given in the following theorem.

**Theorem B.2** (Matrix Bernstein's inequality for U-statistics)**.** Let $k(\cdot) : \mathcal{X} \times \mathcal{X} \to \mathbb{R}^{d\times d}$ be a matrix value function. Let $X_1, \ldots, X_n$ be $n$ independent observations of an random variable $X \in \mathcal{X}$. Suppose that, for any $i \neq i' \in \{1, \ldots, n\}$, $\mathbb{E}k(X_i, X_{i'})$ exists and there exist two constants $R_1, R_2 > 0$ such that

$$\|k(X_i, X_{i'}) - \mathbb{E}k(X_i, X_{i'})\|_2 \leq R_1 \text{ and } \|\mathbb{E}\{k(X_i, X_{i'}) - \mathbb{E}k(X_i, X_{i'})\}^2\|_2 \leq R_2. \tag{B.2}$$

We then have

$$\mathbb{P}\left(\Big\|\frac{1}{\binom{n}{2}}\sum_{i<i'} k(X_i, X_{i'}) - \mathbb{E}k(X_1, X_2)\Big\|_2 \geq t\right) \leq d\exp\Big(-\frac{(n/4)t^2}{R_2 + R_1 t/3}\Big)$$
$$\leq \begin{cases} d \cdot \exp\Big(-\dfrac{3nt^2}{16R_2}\Big), & \text{for } t \leq R_2/R_1; \\ d \cdot \exp\Big(-\dfrac{3nt}{16R_1}\Big), & \text{for } t > R_2/R_1. \end{cases}$$

*Proof.* The proof is the combination of the Hoeffding's decoupling trick and the proof of the independent matrix Bernstein's inequality shown in Tropp (2012). A detailed analysis is given in the proofs of Theorem 2.1 in Wegkamp and Zhao (2016) and Theorem 3.1 in Han and Liu (2016). We refer to theirs for details. □

With the matrix Bernstein inequality of U-statistics, we proceed to prove Theorem 3.1. This is equivalent to calculating $R_1$ and $R_2$ in (B.2) for the particular U-statistics $k_{\mathsf{MK}}$ defined in (2.5).

*Proof of Theorem 3.1.* Let's first calculate the terms $R_1$ and $R_2$ in Theorem B.2 for the particular kernel function

$$k_{\mathsf{MK}}(\boldsymbol{X}_i, \boldsymbol{X}_{i'}) := \frac{(\boldsymbol{X}_i - \boldsymbol{X}_{i'})(\boldsymbol{X}_i - \boldsymbol{X}_{i'})^T}{\|\boldsymbol{X}_i - \boldsymbol{X}_{i'}\|_2^2}.$$

First, we have

$$\Big\|\frac{(\boldsymbol{X}_i - \boldsymbol{X}_{i'})(\boldsymbol{X}_i - \boldsymbol{X}_{i'})^T}{\|\boldsymbol{X}_i - \boldsymbol{X}_{i'}\|_2^2} - \mathbf{K}\Big\|_2 \leq \Big\|\frac{(\boldsymbol{X}_i - \boldsymbol{X}_{i'})(\boldsymbol{X}_i - \boldsymbol{X}_{i'})^T}{\|\boldsymbol{X}_i - \boldsymbol{X}_{i'}\|_2^2}\Big\|_2 + \|\mathbf{K}\|_2 = 1 + \|\mathbf{K}\|_2,$$

where in the last equality we use the fact that

$$\Big\|\frac{(\boldsymbol{X}_i - \boldsymbol{X}_{i'})(\boldsymbol{X}_i - \boldsymbol{X}_{i'})^T}{\|\boldsymbol{X}_i - \boldsymbol{X}_{i'}\|_2^2}\Big\|_2 = \mathrm{Tr}\Big(\frac{(\boldsymbol{X}_i - \boldsymbol{X}_{i'})(\boldsymbol{X}_i - \boldsymbol{X}_{i'})^T}{\|\boldsymbol{X}_i - \boldsymbol{X}_{i'}\|_2^2}\Big) = 1.$$

Secondly, by simple algebra, we have

$$\|\mathbb{E}\{k_{\mathsf{MK}}(\boldsymbol{X}_i, \boldsymbol{X}_{i'}) - \mathbb{E}k_{\mathsf{MK}}(\boldsymbol{X}_i, \boldsymbol{X}_{i'})\}^2\|_2 \leq \|\mathbf{K}\|_2 + \|\mathbf{K}\|_2^2.$$

Accordingly, applying $R_1 = 1 + \|\mathbf{K}\|_2$ and $R_2 = \|\mathbf{K}\|_2 + \|\mathbf{K}\|_2^2$ to Theorem B.2, we have, for any small enough $t$ such that $t \leq (\|\mathbf{K}\|_2 + \|\mathbf{K}\|_2^2)/(1 + \|\mathbf{K}\|_2) = \|\mathbf{K}\|_2$,

$$\mathbb{P}\Big(\Big\|\frac{1}{\binom{n}{2}}\sum_{i<i'} \frac{(\boldsymbol{X}_i - \boldsymbol{X}_{i'})(\boldsymbol{X}_i - \boldsymbol{X}_{i'})^T}{\|\boldsymbol{X}_i - \boldsymbol{X}_{i'}\|_2^2} - \mathbf{K}\Big\|_2 \geq t\Big) \leq d\exp\Big(-\frac{3nt^2}{16(\|\mathbf{K}\|_2 + \|\mathbf{K}\|_2^2)}\Big).$$

Setting

$$t = \sqrt{\frac{16}{3} \cdot \frac{(\|\mathbf{K}\|_2 + \|\mathbf{K}\|_2^2)(\log d + \log(1/\alpha))}{n}} = \|\mathbf{K}\|_2 \sqrt{\frac{16}{3} \cdot \frac{(1 + r^*(\mathbf{K}))(\log d + \log(1/\alpha))}{n}},$$

we get the desired concentration result. □

We then proceed to the proofs of Theorem 3.5 and Corollary 3.2, which exploit the results in Proposition 2.1 and the concentration inequality of the quadratic terms of the Gaussian distribution.

*Proof of Theorem 3.5.* Using Proposition 2.1, the population multivariate Kendall's tau statistic $\mathbf{K}$ has, for $j = 1, \ldots, d$,

$$\lambda_j(\mathbf{K}) = \mathbb{E}\Big(\frac{\lambda_j(\mathbf{\Sigma})Y_j^2}{\lambda_1(\mathbf{\Sigma})Y_1^2 + \cdots + \lambda_d(\mathbf{\Sigma})Y_d^2}\Big) = \mathbb{E}\Big(\frac{Z_j^2}{\sum_{i=1}^d Z_i^2}\Big),$$

where $(Z_1, \ldots, Z_d)^T \sim N_d(\mathbf{0}, \mathbf{\Lambda})$. Here $\mathbf{\Lambda}$ is a diagonal matrix with $\mathbf{\Lambda}_{jj} = \lambda_j(\mathbf{\Sigma})$. Using Lemma B.8 and the fact that $0 \le Z_j^2/(Z_1^2 + \cdots + Z_d^2) \le 1$, by setting $t = 4\log d$, $A = Z_j^2/\sum_{i=1}^d Z_i^2$, and recalling $\mathbb{I}(\cdot)$ to be the indicator function, we have

$$\begin{aligned}
\mathbb{E}A &= \mathbb{E}\Big(A\, \mathbb{I}\Big(\sum_{i=1}^d Z_i^2 \ge \mathrm{Tr}(\mathbf{\Sigma}) - 4\|\mathbf{\Sigma}\|_{\mathsf{F}}\sqrt{\log d}\Big)\Big) + \mathbb{E}\Big(A\, \mathbb{I}\Big(\sum_{i=1}^d Z_i^2 < \mathrm{Tr}(\mathbf{\Sigma}) - 4\|\mathbf{\Sigma}\|_{\mathsf{F}}\sqrt{\log d}\Big)\Big) \\
&\le \frac{\lambda_j(\mathbf{\Sigma})}{\mathrm{Tr}(\mathbf{\Sigma}) - 4\|\mathbf{\Sigma}\|_{\mathsf{F}}\sqrt{\log d}} + \mathbb{P}\Big(\sum_{i=1}^d Z_i^2 < \mathrm{Tr}(\mathbf{\Sigma}) - 4\|\mathbf{\Sigma}\|_{\mathsf{F}}\sqrt{\log d}\Big) \\
&\le \frac{\lambda_j(\mathbf{\Sigma})}{\mathrm{Tr}(\mathbf{\Sigma}) - 4\|\mathbf{\Sigma}\|_{\mathsf{F}}\sqrt{\log d}} + \frac{1}{d^4}.
\end{aligned}$$

Similarly, we have

$$\begin{aligned}
\mathbb{E}A =& \mathbb{E}\Big(A\, \mathbb{I}\Big(\sum_{i=1}^d Z_i^2 \le \mathrm{Tr}(\mathbf{\Sigma}) + 4\|\mathbf{\Sigma}\|_{\mathsf{F}}\sqrt{\log d} + 8\|\mathbf{\Sigma}\|_2 \log d\Big)\Big) \\
&+ \mathbb{E}\Big(A\, \mathbb{I}\Big(\sum_{i=1}^d Z_i^2 > \mathrm{Tr}(\mathbf{\Sigma}) + 4\|\mathbf{\Sigma}\|_{\mathsf{F}}\sqrt{\log d} + 8\|\mathbf{\Sigma}\|_2 \log d\Big)\Big) \\
\ge& \frac{\lambda_j(\mathbf{\Sigma})}{\mathrm{Tr}(\mathbf{\Sigma}) + 4\|\mathbf{\Sigma}\|_{\mathsf{F}}\sqrt{\log d} + 8\|\mathbf{\Sigma}\|_2 \log d} - \frac{\mathbb{E}Z_j^2\, \mathbb{I}\Big(\sum_{i=1}^d Z_i^2 > \mathrm{Tr}(\mathbf{\Sigma}) + 4\|\mathbf{\Sigma}\|_{\mathsf{F}}\sqrt{\log d} + 8\|\mathbf{\Sigma}\|_2 \log d\Big)}{\mathrm{Tr}(\mathbf{\Sigma}) + 4\|\mathbf{\Sigma}\|_{\mathsf{F}}\sqrt{\log d} + 8\|\mathbf{\Sigma}\|_2 \log d}.
\end{aligned}$$

For the above second term, by Cauchy-Swartz inequality, we have

$$\begin{aligned}
&\mathbb{E}Z_j^2\, \mathbb{I}\Big(\sum_{i=1}^d Z_i^2 > \mathrm{Tr}(\mathbf{\Sigma}) + 4\|\mathbf{\Sigma}\|_{\mathsf{F}}\sqrt{\log d} + 8\|\mathbf{\Sigma}\|_2 \log d\Big) \\
\le& (\mathbb{E}Z_j^4)^{1/2} \cdot \Big(\mathbb{P}\Big(\sum_{i=1}^d Z_i^2 > \mathrm{Tr}(\mathbf{\Sigma}) + 4\|\mathbf{\Sigma}\|_{\mathsf{F}}\sqrt{\log d} + 8\|\mathbf{\Sigma}\|_2 \log d\Big)\Big)^{1/2} \\
\le& \sqrt{3}\lambda_j(\mathbf{\Sigma}) \cdot d^{-2}.
\end{aligned}$$

This completes the proof. □

*Proof of Corollary 3.2.* Noticing that

$$\|\mathbf{K}\|_{\max} = \mathbb{E}\frac{(X_j - \mu_j)^2}{\sum_{i=1}^d (X_i - \mu_i)^2}$$

where $\mathbb{E}(X_j - \mu_j)^2 = \|\mathbf{\Sigma}\|_{\max}$ and $\sum_{i=1}^d (X_i - \mu_i)^2 \stackrel{d}{=} \sum_{i=1}^d Z_i^2$, the proof is a line-by-line follow of that of Theorem 3.5. □

## B.3 Proofs of Results in Section 4

In this section we provide the proof of Theorems 4.1 and 4.2.

### B.3.1 Proof of Theorem 4.1

The proof of Theorem 4.1 is via combining the following two lemmas.

**Lemma B.3.** Remind that $S(\boldsymbol{X})$ is the self-normalized version of $\boldsymbol{X}$ defined in (4.4) and $\mathbf{K} := \mathbb{E}S(\boldsymbol{X})S(\boldsymbol{X})^T$ is the multivariate Kendall's tau statistic. Suppose that $\boldsymbol{X} \sim EC_d(\boldsymbol{\mu}, \mathbf{\Sigma}, \xi)$ is elliptically distributed. For any $\boldsymbol{v} \in \mathbb{S}^{d-1}$, suppose that

$$\mathbb{E}\exp\left(t\left[(\boldsymbol{v}^T S(\boldsymbol{X}))^2 - \boldsymbol{v}^T \mathbf{K} \boldsymbol{v}\right]\right) \le \exp(\eta t^2), \quad \text{for } t \le c_0/\sqrt{\eta}, \tag{B.3}$$

where $\eta > 0$ only depends on the eigenvalues of $\mathbf{\Sigma}$ and $c_0$ is an absolute constant. We then have, with probability no smaller than $1 - 2\alpha$, for large enough $n$,

$$\sup_{\boldsymbol{v} \in \mathbb{S}^{d-1} \cap \mathbb{B}_0(s)} \left| \boldsymbol{v}^T \left(\widehat{\mathbf{K}} - \mathbf{K}\right) \boldsymbol{v} \right| \le 2(8\eta)^{1/2} \sqrt{\frac{s(3 + \log(d/s)) + \log(1/\alpha)}{n}}. \tag{B.4}$$

*Proof.* This is a standard argument for sparse PCA, combined with the Hoefdding's decoupling trick. We defer the proof to the last section. □

The next lemma calculates the exact value of $\eta$ in Equation (B.3).

**Lemma B.4.** For any $\boldsymbol{v} = (v_1, \ldots, v_d)^T \in \mathbb{S}^{d-1}$, Equation (B.3) holds with

$$\eta = \sup_{\boldsymbol{v} \in \mathbb{S}^{d-1}} 2\|\boldsymbol{v}^T S(\boldsymbol{X})\|_{\psi_2}^2 + \|\mathbf{K}\|_2$$

and

$$\sup_{\boldsymbol{v} \in \mathbb{S}^{d-1}} \left\|\boldsymbol{v}^T S(\boldsymbol{X})\right\|_{\psi_2} = \sup_{\boldsymbol{v} \in \mathbb{S}^{d-1}} \left\| \frac{\sum_{i=1}^d v_i \lambda_i^{1/2}(\mathbf{\Sigma}) Y_i}{\sqrt{\sum_{i=1}^d \lambda_i(\mathbf{\Sigma}) Y_i^2}} \right\|_{\psi_2},$$

where $(Y_1, \ldots, Y_d)^T \sim N_d(\mathbf{0}, \mathbf{I}_d)$ is standard Gaussian.

*Proof.* The first assertion is a simple consequence of the relationship between the subgaussian and sub-exponential distributions and the property of the sub-exponential distribution (check, for example, Section 5.2 in Vershynin (2010)).

We then focus on the second assertion. Remind that $S(\boldsymbol{X})$ is defined as:

$$S(\boldsymbol{X}) = \frac{\boldsymbol{X} - \widetilde{\boldsymbol{X}}}{\|\boldsymbol{X} - \widetilde{\boldsymbol{X}}\|_2} \overset{\mathsf{d}}{=} \frac{\boldsymbol{X}^*}{\|\boldsymbol{X}^*\|_2} \overset{\mathsf{d}}{=} \frac{\boldsymbol{Z}^0}{\|\boldsymbol{Z}^0\|_2},$$

where $\boldsymbol{X}^* \sim EC_d(\mathbf{0}, \boldsymbol{\Sigma}, \xi^*)$ for some random variable $\xi^* \geq 0$ with $\mathbb{P}(\xi^* = 0) = 0$ and $\boldsymbol{Z}^0 \sim N_d(\mathbf{0}, \boldsymbol{\Sigma})$. Here the second equality is due to the fact that the summation of two independently and identically distributed elliptical random vectors are elliptical distributed (see, for example, Lemma 1 in Lindskog et al. (2003) for a proof). The third equality holds because $\boldsymbol{X}^* \overset{\mathsf{d}}{=} \xi' \boldsymbol{Z}^0$ for some random variable $\xi' \geq 0$ with $\mathbb{P}(\xi' = 0) = 0$. Accordingly, we have $\boldsymbol{Z}^0 / \|\boldsymbol{Z}^0\|_2 \overset{\mathsf{d}}{=} S(\boldsymbol{X})$.

We write $\boldsymbol{\Sigma} = \mathbf{U}\boldsymbol{\Lambda}\mathbf{U}^T$ to be the singular value decomposition of $\boldsymbol{\Sigma}$, where $\boldsymbol{\Lambda}$ has diagonal entries $\lambda_1(\boldsymbol{\Sigma}), \ldots, \lambda_d(\boldsymbol{\Sigma})$. Letting $\boldsymbol{Z} = (Z_1, \ldots, Z_d)^T \sim N_d(\mathbf{0}, \boldsymbol{\Lambda})$ be Gaussian distributed, we have $\boldsymbol{Z}^0 \overset{\mathsf{d}}{=} \mathbf{U}\boldsymbol{Z}$ and can continue to write

$$\frac{\boldsymbol{v}^T \boldsymbol{Z}^0}{\|\boldsymbol{Z}^0\|_2} \overset{\mathsf{d}}{=} \frac{\boldsymbol{w}^T \boldsymbol{Z}}{\|\boldsymbol{Z}\|_2} \overset{\mathsf{d}}{=} \frac{\sum_{i=1}^d w_i \lambda_i^{1/2}(\boldsymbol{\Sigma}) Y_i}{\sqrt{\sum_{i=1}^d \lambda_i(\boldsymbol{\Sigma}) Y_i^2}}, \tag{B.5}$$

where $\boldsymbol{w} = (w_1, \ldots, w_d)^T = \mathbf{U}^T \boldsymbol{v}$, $\boldsymbol{Y} = (Y_1, \ldots, Y_d)^T \sim N_d(\mathbf{0}, \mathbf{I}_d)$, and we have $\boldsymbol{w}^T \boldsymbol{w} = 1$. Because $\mathbf{U}$ is full rank, there is a one to one map between $\boldsymbol{v}$ and $\boldsymbol{w}$, and hence taking supremum over $\boldsymbol{v}$ is equivalent to taking supremum over $\boldsymbol{w}$. This completes the proof. □

### B.3.2 Proof of Theorem 4.2

In this section we focus on the proof of Theorem 4.2. We aim at providing sharp subgaussian constant of $S(\boldsymbol{X})$.

*Proof of Theorem 4.2.* For any $\boldsymbol{v} \in \mathbb{S}^{d-1}$, it is enough to show that $\boldsymbol{v}^T S(\boldsymbol{X}) \in \mathbb{R}$ is subgaussian distributed with subgaussian norm uniformly bounded by $\sqrt{2\lambda_1(\boldsymbol{\Sigma})/(q\lambda_q(\boldsymbol{\Sigma}))}$. For notational simplicity, with an abuse of notation, we let $\lambda_i(\boldsymbol{\Sigma})$ be abbreviated as $\lambda_i$ for $i = 1, \ldots, q$.

For any $p = 1, 2, \ldots$ and $\boldsymbol{v} \in \mathbb{S}^{d-1}$, to show $\boldsymbol{v}^T S(\boldsymbol{X})$ is subgaussian, following (B.5) in the proof of Theorem 4.1, it is sufficient to bound its all higher moments:

$$\mathbb{E}\Big(\Big|\frac{\boldsymbol{v}^T \boldsymbol{Z}^0}{\|\boldsymbol{Z}^0\|_2}\Big|\Big)^p = \mathbb{E}\Bigg(\Bigg|\frac{\sum_{i=1}^q w_i \lambda_i^{1/2} Y_i}{\sqrt{\sum_{i=1}^q \lambda_i Y_i^2}}\Bigg|\Bigg)^p \leq \mathbb{E}\Bigg(\Bigg|\frac{\sum_{i=1}^q w_i (\lambda_i/\lambda_q)^{1/2} Y_i}{\sqrt{\sum_{i=1}^q Y_i^2}}\Bigg|\Bigg)^p, \tag{B.6}$$

where $\boldsymbol{w} = (w_1, \ldots, w_d)^T = \mathbf{U}^T \boldsymbol{v}$, $\boldsymbol{Y} = (Y_1, \ldots, Y_q)^T \sim N_q(\mathbf{0}, \mathbf{I}_q)$, and we have $\boldsymbol{w}^T \boldsymbol{w} = 1$.

Next we prove that the rightest term at (B.6) reaches its maximum at $\boldsymbol{w}^* = (1, 0, \ldots, 0)^T$. To this end, we adopt a standard technique in calculating the distribution of the quadratic ratio (see, for example, Provost and Cheong (2000)). Let $\boldsymbol{\zeta}(\boldsymbol{w}) = (\zeta_1, \ldots, \zeta_q)$ with $\zeta_i = w_i(\lambda_i/\lambda_q)^{1/2}$. For any constant $c \geq 0$, we have

$$\mathbb{P}\Big(\Big|\frac{\boldsymbol{\zeta}(\boldsymbol{w})^T \boldsymbol{Y}}{\|\boldsymbol{Y}\|_2}\Big| > c\Big) = \mathbb{P}\Big(\frac{\boldsymbol{Y}^T \boldsymbol{\zeta}(\boldsymbol{w}) \boldsymbol{\zeta}(\boldsymbol{w})^T \boldsymbol{Y}}{\boldsymbol{Y}^T \boldsymbol{Y}} > c^2\Big) = \mathbb{P}\big(\boldsymbol{Y}^T (\boldsymbol{\zeta}(\boldsymbol{w})\boldsymbol{\zeta}(\boldsymbol{w})^T - c^2 \mathbf{I}_d) \boldsymbol{Y} > 0\big). \tag{B.7}$$

Then it is immediate to have

$$\mathbb{P}\big(\boldsymbol{Y}^T(\boldsymbol{\zeta}(\boldsymbol{w})\boldsymbol{\zeta}(\boldsymbol{w})^T - c^2\mathbf{I}_d)\boldsymbol{Y} > 0\big) = \mathbb{P}\Big(\sum_{j=1}^{q} l_j Y_j^2 > 0\Big),$$

where $l_1 = \boldsymbol{\zeta}(\boldsymbol{w})^T\boldsymbol{\zeta}(\boldsymbol{w}) - c^2$ and $l_j = -c^2$ for $j = 2, \ldots, q$. This implies that

$$\operatorname*{arg\,max}_{\boldsymbol{w}\in\mathbb{S}^{d-1}} \mathbb{P}\left(\boldsymbol{Y}^T(\boldsymbol{\zeta}(\boldsymbol{w})\boldsymbol{\zeta}(\boldsymbol{w})^T - c^2\mathbf{I}_d)\boldsymbol{Y} > 0\right) = \operatorname*{arg\,max}_{\boldsymbol{w}\in\mathbb{S}^{d-1}} \boldsymbol{\zeta}(\boldsymbol{w})^T\boldsymbol{\zeta}(\boldsymbol{w}) = (1, 0, \ldots, 0)^T = \boldsymbol{w}^*.$$

In other words, for any $c > 0$ and $\boldsymbol{w} \in \mathbb{S}^{d-1}$, we have

$$\mathbb{P}\Big(\Big|\frac{\boldsymbol{\zeta}(\boldsymbol{w}^*)^T\boldsymbol{Y}}{\|\boldsymbol{Y}\|_2}\Big| > c\Big) \geq \mathbb{P}\Big(\Big|\frac{\boldsymbol{\zeta}(\boldsymbol{w})^T\boldsymbol{Y}}{\|\boldsymbol{Y}\|_2}\Big| > c\Big).$$

Then (B.6) further implies that

$$\sup_{\boldsymbol{v}\in\mathbb{S}^{d-1}} \mathbb{E}\Big(\Big|\frac{\boldsymbol{v}^T\boldsymbol{Z}^0}{\|\boldsymbol{Z}^0\|_2}\Big|\Big)^p \leq \mathbb{E}\left(\left|\frac{(\lambda_1/\lambda_q)^{1/2}Y_i}{\sqrt{\sum_{i=1}^q Y_1^2}}\right|\right)^p = (\lambda_1/\lambda_q)^{p/2} \cdot \mathbb{E}\left(\frac{Y_1}{\sqrt{\sum_{i=1}^q Y_i^2}}\right)^p. \tag{B.8}$$

In the end, combining (B.8) and Lemma B.9, we have

$$\sup_{\boldsymbol{v}\in\mathbb{S}^{d-1}} \big\|\boldsymbol{v}^T S(\boldsymbol{X})\big\|_{\psi_2} = \sup_{\boldsymbol{v}\in\mathbb{S}^{d-1}} \Big\|\frac{\boldsymbol{v}^T\boldsymbol{Z}^0}{\|\boldsymbol{Z}^0\|_2}\Big\|_{\psi_2} \leq \sqrt{\frac{\lambda_1}{\lambda_q} \cdot \frac{2}{q}}.$$

This completes the proof. $\square$

### B.4 Proofs of Results in Section 5

In this section we provide the proofs of Theorems 5.2, 5.3, and 5.4.

#### B.4.1 Proof of Theorem 5.2

Theorem 5.2 is the direct consequence of the following lemma by setting $\mathbf{A} = \mathbf{X}_m$, $\mathbf{C} = \widehat{\mathbf{X}}_m$, and $\mathbf{P} = \mathbf{\Pi}_m$.

**Lemma B.5.** For any positive semidefinite symmetric matrix $\mathbf{A} \in \mathbb{R}^{d\times d}$ (not necessarily satisfying $\mathbf{A} \preceq \mathbf{I}_d$) and rank $m$ projection matrix $\mathbf{P}$, letting $\mathbf{C} = \sum_{j=1}^m \boldsymbol{u}_j(\mathbf{A})\boldsymbol{u}_j(\mathbf{A})^T$, we have

$$\|\mathbf{C} - \mathbf{P}\|_{\mathsf{F}} \leq 4\|\mathbf{A} - \mathbf{P}\|_{\mathsf{F}}.$$

*Proof.* We let $\epsilon := \|\mathbf{A} - \mathbf{P}\|_{\mathsf{F}}$. We first define $\mathbf{B} := \sum_{j=1}^m \lambda_j(\mathbf{A})\boldsymbol{u}_j(\mathbf{A})\boldsymbol{u}_j(\mathbf{A})^T$ to be the best rank $m$ approximation to $\mathbf{A}$. By simple algebra, we have

$$\|\mathbf{A} - \mathbf{B}\|_{\mathsf{F}}^2 = \sum_{j>m}(\lambda_j(\mathbf{A}))^2,$$

and accordingly, using triangular inequality,

$$\|\mathbf{B}-\mathbf{P}\|_{\mathsf{F}} \le \|\mathbf{A}-\mathbf{B}\|_{\mathsf{F}} + \|\mathbf{A}-\mathbf{P}\|_{\mathsf{F}} = \Big(\sum_{j>m}(\lambda_j(\mathbf{A}))^2\Big)^{1/2} + \epsilon.$$

Using Lemma B.10 and the fact that $\lambda_j(\mathbf{P}) = 0$ for all $j > m$, we further have

$$\sum_{j>m}(\lambda_j(\mathbf{A}))^2 \le \sum_{j=1}^{d}(\lambda_j(\mathbf{A}) - \lambda_j(\mathbf{P}))^2 \le \|\mathbf{A}-\mathbf{P}\|_{\mathsf{F}}^2 = \epsilon^2,$$

so that $\|\mathbf{B}-\mathbf{P}\|_{\mathsf{F}} \le 2\epsilon$. With a little abuse of notation, for $j = 1, \ldots, d$, we write $\lambda_j = \lambda_j(\mathbf{A})$ and $\boldsymbol{u}_j = \boldsymbol{u}_j(\mathbf{A})$ for simplicity. Therefore, we have

$$\|\mathbf{B}-\mathbf{P}\|_{\mathsf{F}}^2 = \Big\langle \sum_{j=1}^{m}\lambda_j\boldsymbol{u}_j\boldsymbol{u}_j^T - \mathbf{P}, \sum_{j=1}^{m}\lambda_j\boldsymbol{u}_j\boldsymbol{u}_j^T - \mathbf{P}\Big\rangle = \sum_{j=1}^{m}\lambda_j^2 + m - 2\sum_{j=1}^{m}\lambda_j\boldsymbol{u}_j^T\mathbf{P}\boldsymbol{u}_j \le 4\epsilon^2.$$

This further implies that

$$\sum_{j=1}^{m}(\lambda_j^2+1) - 4\epsilon^2 \le 2\sum_{j=1}^{m}\lambda_j\boldsymbol{u}_j^T\mathbf{P}\boldsymbol{u}_j \le 2\sum_{j=1}^{m}\lambda_j \Rightarrow \sum_{j=1}^{m}(1-\lambda_j)^2 \le 4\epsilon^2.$$

Noticing that, by the definition of $\mathbf{B}$,

$$\|\mathbf{C}-\mathbf{B}\|_{\mathsf{F}}^2 = \sum_{j=1}^{m}(1-\lambda_j)^2 \le 4\epsilon^2,$$

we finally have

$$\|\mathbf{C}-\mathbf{P}\|_{\mathsf{F}} \le \|\mathbf{B}-\mathbf{P}\|_{\mathsf{F}} + \|\mathbf{C}-\mathbf{B}\|_{\mathsf{F}} \le 4\epsilon.$$

This completes the proof. □

### B.4.2 Proof of Theorem 5.3

To prove Theorem 5.3, we first provide a general theorem, which quantifies the convergence rate of a U-statistic estimate of the covariance matrix.

**Theorem B.6** (Concentration inequality for U-statistics estimators of covariance matrix)**.** Let $k_1(\cdot) : \mathcal{X} \times \mathcal{X} \to \mathbb{R}$ and $k_2(\cdot) : \mathcal{X} \times \mathcal{X} \to \mathbb{R}$ be two real functions and $X_1, \ldots, X_n$ be $n$ observations of the random variable $X \in \mathcal{X}$ satisfying $\|k_i(X_1, X_2)\|_{\psi_2} \le K_0$ for $i = 1, 2$, and $\tau^2 := \mathbb{E}k_1(X_1, X_2)k_2(X_1, X_2)$. We then have,

$$\mathbb{P}\Big(\frac{1}{\binom{n}{2}}\sum_{i<i'}k_1(X_i, X_{i'})k_2(X_i, X_{i'}) - \mathbb{E}k_1(X_i, X_{i'})k_2(X_i, X_{i'}) \ge t\Big) \le \exp\Big(-\frac{nt^2}{8C(4K_0^2+\tau^2)^2}\Big),$$

for all $t < 2Cc(4K_0^2 + \tau^2)$ and some generic constants $C$ and $c$. When the two kernel functions $k_1(\cdot)$ and $k_2(\cdot)$ are equal, the term $4K_0^2$ above can be further relaxed to be $2K_0^2$.

*Proof.* First, let's calculate $\mathbb{E}\exp(t\cdot\frac{1}{\binom{n}{2}}\sum_{i<i'}k_1(X_i,X_{i'})k_2(X_i,X_{i'})-\tau^2)$. Using a similar decoupling technique as in Theorem B.2, we have

$$\mathbb{E}\exp\Big(\frac{t}{\binom{n}{2}}\sum_{i<i'}k_1(X_i,X_{i'})k_2(X_i,X_{i'})-\mathbb{E}k_1(X_i,X_{i'})k_2(X_i,X_{i'})\Big)$$
$$=\Big(\mathbb{E}e^{\frac{t}{m}(k_1(X_1,X_2)k_2(X_1,X_2)-\mathbb{E}k_1(X_1,X_2)k_2(X_1,X_2))}\Big)^m,$$

where $m:=n/2$. We then have

$$\|k_1(X_1,X_2)k_2(X_1,X_2)\|_{\psi_1}=\Big\|\frac{1}{4}\{(k_1(X_1,X_2)+k_2(X_1,X_2))^2-(k_1(X_1,X_2)-k_2(X_1,X_2))^2\}\Big\|_{\psi_1}$$
$$\leq\frac{1}{4}\left\|(k_1(X_1,X_2)+k_2(X_1,X_2))^2\right\|_{\psi_1}+\frac{1}{4}\left\|(k_1(X_1,X_2)-k_2(X_1,X_2))^2\right\|_{\psi_1}. \quad \text{(B.9)}$$

Using Minkowski's inequality, for any two random variables $X,Y\in\mathbb{R}$, we have $(\mathbb{E}(X+Y)^p)^{1/p}\leq(\mathbb{E}X^p)^{1/p}+(\mathbb{E}Y^p)^{1/p}$. Accordingly, we have $\|X+Y\|_{\psi_2}\leq\|X\|_{\psi_2}+\|Y\|_{\psi_2}$. This implies that

$$\|k_1(X_1,X_2)\pm k_2(X_1,X_2)\|_{\psi_2}\leq\|k_1(X_1,X_2)\|_{\psi_2}+\|k_2(X_1,X_2)\|_{\psi_2}\leq 2K_0.$$

Therefore, using the relationship between $\|\cdot\|_{\psi_1}$ and $\|\cdot\|_{\psi_2}$, we have

$$\left\|(k_1(X_1,X_2)\pm k_2(X_1,X_2))^2\right\|_{\psi_1}\leq 2(\|k_1(X_1,X_2)\pm k_2(X_1,X_2)\|_{\psi_2})^2\leq 8K_0^2.$$

This, combined with (B.9), implies that

$$\|k_1(X_1,X_2)k_2(X_1,X_2)\|_{\psi_1}\leq 4K_0^2. \quad \text{(B.10)}$$

Accordingly, $k_1(X_1,X_2)k_2(X_1,X_2)-\mathbb{E}k_1(X_1,X_2)k_2(X_1,X_2)$ is sub-exponential and has sub-exponential norm

$$\|k_1(X_1,X_2)k_2(X_1,X_2)-\mathbb{E}k_1(X_1,X_2)k_2(X_1,X_2)\|_{\psi_1}\leq 4K_0^2+\tau^2.$$

We can then apply Lemma 5.15 in Vershynin (2010) to deduce that

$$\mathbb{E}\exp\Big(\frac{t}{\binom{n}{2}}\sum_{i<i'}k_1(X_i,X_{i'})k_2(X_i,X_{i'})-\tau^2\Big)\leq\exp\Big(\frac{C(4K_0^2+\tau^2)^2t^2}{m}\Big),\ \text{for}\ \Big|\frac{t}{m}\Big|\leq c/(4K_0^2+\tau^2),$$

where $C$ and $c$ are two absolute constants. We then use the Markov's inequality to have the final concentration inequality. This completes the proof of the first part.

Furthermore, when $k_1(\cdot)=k_2(\cdot)$, we can improve the upper bound in (B.10) to be $2K_0^2$. And the whole proof still proceeds. This completes the proof of the second part. □

Using Theorem B.6, we are now ready to prove Theorem 5.3.

*Proof of Theorem 5.3.* Using the result in Theorem 4.2, we have for any $\boldsymbol{v}\in\mathbb{S}^{d-1}$,

$$\|\boldsymbol{v}^TS(\boldsymbol{X})\|_{\psi_2}\leq\sqrt{\frac{\lambda_1(\boldsymbol{\Sigma})}{\lambda_q(\boldsymbol{\Sigma})}\cdot\frac{2}{q}}.$$

In particular, for any $j, k \in \{1, \ldots, d\}$, setting

$$k_1(\boldsymbol{X}_1, \boldsymbol{X}_2) = \frac{\boldsymbol{e}_j^T(\boldsymbol{X}_1 - \boldsymbol{X}_2)}{\|\boldsymbol{X}_1 - \boldsymbol{X}_2\|_2} \quad \text{and} \quad k_2(\boldsymbol{X}_1, \boldsymbol{X}_2) := \frac{\boldsymbol{e}_k^T(\boldsymbol{X}_1 - \boldsymbol{X}_2)}{\|\boldsymbol{X}_1 - \boldsymbol{X}_2\|_2},$$

we have

$$\|k_i(\boldsymbol{X}_1, \boldsymbol{X}_2)\|_{\psi_2} \leq \sqrt{\frac{\lambda_1(\boldsymbol{\Sigma})}{\lambda_q(\boldsymbol{\Sigma})} \cdot \frac{2}{q}}, \quad \text{for } i = 1, 2.$$

Accordingly, using Theorem B.6, we have

$$\mathbb{P}(|\widehat{\mathbf{K}}_{jk} - \mathbf{K}_{jk}| \geq t) \leq \exp\Big(-\frac{nt^2}{8C(8(\lambda_1(\boldsymbol{\Sigma})/q\lambda_q(\boldsymbol{\Sigma})) + \mathbf{K}_{jk})^2}\Big), \quad \text{for } t < 2Cc(8\lambda_1(\boldsymbol{\Sigma})/q\lambda_q(\boldsymbol{\Sigma}) + \mathbf{K}_{jk}).$$

We then use the union bound to deduce that

$$\mathbb{P}(\|\widehat{\mathbf{K}} - \mathbf{K}\|_{\max} \geq t) \leq d^2 \exp\Big(-\frac{nt^2}{8C(8\lambda_1(\boldsymbol{\Sigma})/q\lambda_q(\boldsymbol{\Sigma}) + \|\mathbf{K}\|_{\max})^2}\Big),$$

which implies that, for large enough $n$, with probability larger than $1 - \alpha^2$,

$$\|\widehat{\mathbf{K}} - \mathbf{K}\|_{\max} \leq 4\sqrt{C}\Big(\frac{8\lambda_1(\boldsymbol{\Sigma})}{q\lambda_q(\boldsymbol{\Sigma})} + \|\mathbf{K}\|_{\max}\Big)\sqrt{\frac{\log d + \log(1/\alpha)}{n}}.$$

This completes the proof. □

### B.4.3 Proof of Theorem 5.4

*Proof of Theorem 5.4.* Without loss of generality, we assume that $(\boldsymbol{u}_1(\mathbf{X}_1))^T\boldsymbol{u}_1(\mathbf{K}) \geq 0$. Using Corollary 5.1, we have

$$\|\boldsymbol{u}_1(\mathbf{X}_1) - \boldsymbol{u}_1(\mathbf{K})\|_\infty \leq \|\boldsymbol{u}_1(\mathbf{X}_1) - \boldsymbol{u}_1(\mathbf{K})\|_2 = O_P(s\sqrt{\log d/n}).$$

It is then immediate that, with high probability, the support set of $\boldsymbol{u}_1(\mathbf{X}_1)$, denoted as $\widehat{J}$, must include $J_0$ and belong to $J_1 := \{j : |(\boldsymbol{u}_1(\mathbf{K}))_j| = \Omega(s\log d/\sqrt{n})\}$, i.e.,

$$\mathbb{P}(J_0 \subset \widehat{J} \subset J_1) \to 1, \quad \text{when } n \to \infty.$$

Therefore, with high probability, $\|\boldsymbol{v}^{(0)}\|_0 \leq \text{card}(J_1) \leq s$. Moreover, we have, for any $j \in J_0$, because $s\sqrt{\log d/n} = o(s\log d/\sqrt{n})$, we have

$$(\boldsymbol{u}_1(\mathbf{X}_1))_j = (\boldsymbol{u}_1(\mathbf{K}))_j(1 + o_P(1))$$

for $j \in J_0$. Using the above result, we have

$$\boldsymbol{v}^{(0)} = \text{TRC}(\boldsymbol{u}_1(\mathbf{K}), \widehat{J})/\|\,\text{TRC}(\boldsymbol{u}_1(\mathbf{K}), \widehat{J})\|_2 \cdot (1 + o_P(1)).$$

Accordingly, under the condition of Theorem 5.4, we have,

$$(\boldsymbol{v}^{(0)})^T\boldsymbol{u}_1(\mathbf{K}) = \|\,\text{TRC}(\boldsymbol{u}_1(\mathbf{K}), \widehat{J})\|_2(1 + o_P(1)) \geq C_3(1 + o_P(1)),$$

and accordingly is asymptotically lower bounded by absolute constant. The rest can be proved by using Theorem 4 in Yuan and Zhang (2013). □

## B.5 Auxiliary Lemmas

In this section we provide the auxiliary lemmas. The first lemma shows that any elliptical distribution is a random scaled version of the Gaussian.

**Lemma B.7.** Let $\boldsymbol{X} \sim EC_d(\boldsymbol{\mu}, \boldsymbol{\Sigma}, \xi)$ be an elliptical distribution with $\boldsymbol{\Sigma} = \mathbf{A}\mathbf{A}^T$. It takes another stochastic representation:

$$\boldsymbol{X} \stackrel{\mathsf{d}}{=} \boldsymbol{\mu} + \xi \boldsymbol{Z} / \|\mathbf{A}^{\dagger} \boldsymbol{Z}\|_2,$$

where $\boldsymbol{Z} \sim N_d(\mathbf{0}, \boldsymbol{\Sigma})$, $\xi \geq 0$ is independent of $\boldsymbol{Z}/\|\mathbf{A}^{\dagger}\boldsymbol{Z}\|_2$, and $\mathbf{A}^{\dagger}$ is the Moore-Penrose pseudoinverse of $\mathbf{A}$.

*Proof.* Let $\boldsymbol{X} = \boldsymbol{\mu} + \xi \mathbf{A}\boldsymbol{U}$ and $q := \text{rank}(\boldsymbol{\Sigma}) = \text{rank}(\mathbf{A})$ as in (2.1). Let $\boldsymbol{U} = \boldsymbol{\epsilon}/\|\boldsymbol{\epsilon}\|_2$ with a standard normal vector $\boldsymbol{\epsilon}$ in $\mathbb{R}^q$. Note that if $\mathbf{A} = \mathbf{V}_1 \mathbf{D} \mathbf{V}_2^T$ is the singular value decomposition of $\mathbf{A} \in \mathbb{R}^{d \times q}$ with $\mathbf{V}_1 \in \mathbb{R}^{d\times q}$ and $\mathbf{D}, \mathbf{V} \in \mathbb{R}^{q \times q}$, then $\mathbf{A}^{\dagger} = \mathbf{V}_2 \mathbf{D}^{-1} \mathbf{V}_1^T$. Since $\text{rank}(\mathbf{A}) = q$, we have $\mathbf{A}^{\dagger}\mathbf{A} = \mathbf{I}_q$. Accordingly, let $\boldsymbol{Z} = \mathbf{A}\boldsymbol{\epsilon} \sim N(0, \boldsymbol{\Sigma})$. It follows that

$$\boldsymbol{X} - \boldsymbol{\mu} = \xi \mathbf{A}\boldsymbol{U} = \xi \boldsymbol{Z}/\|\boldsymbol{\epsilon}\|_2 = \xi \boldsymbol{Z}/\|\mathbf{A}^{\dagger}\boldsymbol{Z}\|_2.$$

The proof is complete. □

The next lemma gives two Hanson-Wright type inequalities for the quadratic term of the Gaussian distributed random vectors.

**Lemma B.8.** Let $\boldsymbol{Z} \sim N_d(\mathbf{0}, \boldsymbol{\Sigma})$ be a $d$-dimensional Gaussian distributed random vector. Then for every $t \geq 0$, we have

$$\mathbb{P}(\boldsymbol{Z}^T\boldsymbol{Z} - \text{Tr}(\boldsymbol{\Sigma}) \leq -2\|\boldsymbol{\Sigma}\|_{\mathsf{F}}\sqrt{t}) \leq \exp(-t),$$

and

$$\mathbb{P}(\boldsymbol{Z}^T\boldsymbol{Z} - \text{Tr}(\boldsymbol{\Sigma}) \geq 2\|\boldsymbol{\Sigma}\|_{\mathsf{F}}\sqrt{t} + 2\|\boldsymbol{\Sigma}\|_2 t) \leq \exp(-t).$$

*Proof.* Let $\mathbf{U}\boldsymbol{\Lambda}\mathbf{U}^T$ be the SVD decomposition of $\boldsymbol{\Sigma}$. Then letting $\boldsymbol{Y} = (Y_1, \ldots, Y_d)^T \sim N_d(\mathbf{0}, \mathbf{I}_d)$, using the fact that $\mathbf{U}^T\boldsymbol{Y} \stackrel{\mathsf{d}}{=} \boldsymbol{Y}$, we have

$$\boldsymbol{Z}^T\boldsymbol{Z} \stackrel{\mathsf{d}}{=} \boldsymbol{Y}^T\boldsymbol{\Sigma}\boldsymbol{Y} = \boldsymbol{Y}^T\mathbf{U}\boldsymbol{\Lambda}\mathbf{U}^T\boldsymbol{Y} \stackrel{\mathsf{d}}{=} \boldsymbol{Y}^T\boldsymbol{\Lambda}\boldsymbol{Y} = \sum \lambda_j(\boldsymbol{\Sigma})Y_j^2.$$

The rest follows from Lemma 1 in Laurent and Massart (2000). □

The next lemma shows that a simple version of the quadratic ratio under the Gaussian assumption is subgaussian.

**Lemma B.9.** For $\boldsymbol{Y} = (Y_1, \ldots, Y_q)^T \sim N_q(\mathbf{0}, \mathbf{I}_q)$, we have

$$\Big\| \frac{Y_j}{\sqrt{\sum_{i=1}^q Y_i^2}} \Big\|_{\psi_2} \leq \sqrt{\frac{2}{q}}, \quad \text{for } j = 1, \ldots, d,$$

where we remind that the subgaussian norm $\|\cdot\|_{\psi_2}$ is defined in (4.3).

*Proof.* It is known (see, for example, Chapter 3 in Bilodeau and Brenner (1999)) that

$$Y_1^2 / \sum_{i=1}^{q} Y_i^2 \sim \text{Beta}\Big(\frac{1}{2}, \frac{q-1}{2}\Big).$$

Accordingly, using the Jensen's inequality and the property of the beta distribution, we have

$$\mathbb{E}\Big(\frac{Y_1^2}{\sum_{i=1}^q Y_i^2}\Big)^{p/2} \le \Big(\mathbb{E}\Big(\frac{Y_1^2}{\sum_{i=1}^q Y_i^2}\Big)^p\Big)^{1/2} = \Big(\prod_{r=0}^{p-1} \frac{2r+1}{2r+q}\Big)^{1/2},$$

where the last equality is using the moment formula of the beta distribution (check, for example, Page 36 in Gupta and Nadarajah (2004)). When $q$ is even, using the Sterling's inequality, we can continue to write

$$\begin{aligned}\Big(\prod_{r=0}^{p-1} \frac{2r+1}{2r+q}\Big)^{1/2} &= \Big(\frac{(2p-1)!!}{(2p+q-2)!!}\Big)^{1/2} \le \Big(\frac{p!}{(p+(q-2)/2)!}\Big)^{1/2} \le \Big(\frac{p^{p+1/2}e^{(q-2)/2}}{(p+(q-2)/2)^{p+(q-1)/2}}\Big)^{1/2} \\ &= p^{p/2+1/(2p)} \cdot \Big(\frac{1}{(p+(q-2)/2)^p} \cdot \frac{e^{(q-2)/2}}{(p+(q-2)/2)^{(q-1)/2}}\Big)^{1/2} \le p^{p/2}/(p+(q-2)/2)^{p/2}.\end{aligned}$$

Accordingly, we have

$$\Big(\mathbb{E}\Big(\frac{Y_1^2}{\sum Y_i^2}\Big)^{p/2}\Big)^{1/p} \le \sqrt{p} \cdot \sqrt{\frac{1}{q}}.$$

Similarly, when $q$ is odd, we have $(\mathbb{E}(Y_1^2/\sum Y_i^2)^{p/2})^{1/p} \le \sqrt{p} \cdot \kappa_U \sqrt{1/(q-1/2)} \le \sqrt{p} \cdot \sqrt{2/q}$. This completes the proof. ∎

The final lemma states a Wyel type inequality and is well known in the matrix perturbation literature (check, for example, Equation (3.3.32) in Horn and Johnson (1991)).

**Lemma B.10.** For any positive semidefinite symmetric matrices $\mathbf{A}, \mathbf{B} \in \mathbb{R}^{d \times d}$ (so that the eigenvalues and singular values are equal), we have

$$\sum_{i=1}^{d} (\lambda_i(\mathbf{A}) - \lambda_i(\mathbf{B}))^2 \le \|\mathbf{A} - \mathbf{B}\|_{\mathsf{F}}^2.$$

# C The Proof of Lemma B.3

The proof of Lemma B.3 is shown in this section. The idea is to combine the proof of sparse PCA (see, for example, Lounici (2013)) with the Hoeffding's decoupling trick. We present the proof here mainly for completedness.

*Proof of Lemma B.3.* Let $a \in \mathbb{Z}^+$ be an integer no smaller than 1 and $J_a$ be any subset of $\{1, \ldots, d\}$ with cardinality $a$. For any $s$-dimensional sphere $\mathbb{S}^{s-1}$ equipped with Euclidean distance, we let $\mathcal{N}_\epsilon$ be a subset of $\mathbb{S}^{s-1}$ such that for any $\boldsymbol{v} \in \mathbb{S}^{s-1}$, there exists $\boldsymbol{u} \in \mathcal{N}_\epsilon$ subject to $\|\boldsymbol{u} - \boldsymbol{v}\|_2 \le \epsilon$. It is known that the cardinal number of $\mathcal{N}_\epsilon$ has an upper bound: $\text{card}(\mathcal{N}_\epsilon) < \big(1 + \frac{2}{\epsilon}\big)^s$. Let $\mathcal{N}_{1/4}$ be a (1/4)-net of $\mathbb{S}^{s-1}$. We then have card($\mathcal{N}_{1/4}$) is upper bounded by $9^s$. Moreover, for any symmetric matrix $\mathbf{M} \in \mathbb{R}^{s\times s}$, we have

$$\sup_{\boldsymbol{v}\in\mathbb{S}^{s-1}} |\boldsymbol{v}^T\mathbf{M}\boldsymbol{v}| \le \frac{1}{1-2\epsilon}\sup_{\boldsymbol{v}\in\mathcal{N}_\epsilon} |\boldsymbol{v}^T\mathbf{M}\boldsymbol{v}|, \ \text{implying} \ \sup_{\boldsymbol{v}\in\mathbb{S}^{s-1}} |\boldsymbol{v}^T\mathbf{M}\boldsymbol{v}| \le 2\sup_{\boldsymbol{v}\in\mathcal{N}_{1/4}} |\boldsymbol{v}^T\mathbf{M}\boldsymbol{v}|.$$

Let $\beta > 0$ be a quantity defined as $\beta := (8\eta)^{1/2}\sqrt{\frac{s(3+\log(d/s))+\log(1/\alpha)}{n}}$. By the union bound, we have

$$\begin{aligned}
&\mathbb{P}\Big(\sup_{\boldsymbol{b}\in S^{s-1}}\sup_{J_s\subset\{1,\cdots,d\}}\Big|\boldsymbol{b}^T\Big[\widehat{\mathbf{K}}-\mathbf{K}\Big]_{J_s,J_s}\boldsymbol{b}\Big|>2\beta\Big) \le \mathbb{P}\Big(\sup_{\boldsymbol{b}\in \mathcal{N}_{1/4}}\sup_{J_s\subset\{1,\cdots,d\}}\Big|\boldsymbol{b}^T\Big[\widehat{\mathbf{K}}-\mathbf{K}\Big]_{J_s,J_s}\boldsymbol{b}\Big|>\beta\Big)\\
\le&9^s\binom{d}{s}\mathbb{P}\Big(\Big|\boldsymbol{b}^T\Big[\widehat{\mathbf{K}}-\mathbf{K}\Big]_{J_s,J_s}\boldsymbol{b}\Big| > (8\eta)^{1/2}\sqrt{\frac{s(3+\log(d/s))+\log(1/\alpha)}{n}}, \text{ for any } \boldsymbol{b} \text{ and } J_s\Big).
\end{aligned}$$

Thus, if we can show that for any $\boldsymbol{b}\in\mathbb{S}^{s-1}$ and $J_s$, we have

$$\mathbb{P}\Big(\Big|\boldsymbol{b}^T\Big[\widehat{\mathbf{K}}-\mathbf{K}\Big]_{J_s,J_s}\boldsymbol{b}\Big| > t\Big) \le 2e^{-nt^2/(8\eta)}, \tag{C.1}$$

for $\eta$ defined in Equation (B.3). Then, using the bound $\binom{d}{s} < (ed/s)^s$, we have

$$\begin{aligned}
&9^s\binom{d}{s}\mathbb{P}\Big(\Big|\boldsymbol{b}^T\Big[\widehat{\mathbf{K}}-\mathbf{K}\Big]_{J_s,J_s}\boldsymbol{b}\Big| > (8\eta)^{1/2}\sqrt{\frac{s(3+\log(d/s))+\log(1/\alpha)}{n}}, \ \text{ for any } \boldsymbol{b} \text{ and } J\Big)\\
\le&2\exp\{s(1+\log 9-\log(s))+s\log d - s(3+\log d-\log s)-\log(1/\alpha)\} \le 2\alpha.
\end{aligned}$$

It shows that, with probability greater than $1-2\alpha$, the bound in Equation (B.4) holds.

We now show that Equation (C.1) holds. For any $t$, we have

$$\mathbb{E}\exp\Big(t\cdot\boldsymbol{b}^T\Big[\widehat{\mathbf{K}}-\mathbf{K}\Big]_{J_s,J_s}\boldsymbol{b}\Big) = \mathbb{E}\exp\Big(t\cdot\frac{1}{\binom{n}{2}}\sum_{i<i'}\boldsymbol{b}^T\Big(\frac{(\boldsymbol{X}_i-\boldsymbol{X}_{i'})_{J_s}(\boldsymbol{X}_i-\boldsymbol{X}_{i'})_{J_s}^T}{\|\boldsymbol{X}_i-\boldsymbol{X}_{i'}\|_2^2}-\mathbf{K}_{J_s,J_s}\Big)\boldsymbol{b}\Big).$$

Let $S_n$ represent the permutation group of $\{1,\dots,n\}$. For any $\sigma\in S_n$, let $(i_1,\dots,i_n) := \sigma(1,\dots,n)$ represent a permuted series of $\{1,\dots,n\}$ and $O(\sigma) := \{(i_1,i_2),(i_3,i_4),\dots,(i_{n-1},i_n)\}$. In particular, we denote $O(\sigma_0) := \{(1,2),(3,4),\dots,(n-1,n)\}$. By simple calculation,

$$\begin{aligned}
&\mathbb{E}\exp\Big(t\cdot\frac{1}{\binom{n}{2}}\sum_{i<i'}\boldsymbol{b}^T\Big(\frac{(\boldsymbol{X}_i-\boldsymbol{X}_{i'})_{J_s}(\boldsymbol{X}_i-\boldsymbol{X}_{i'})_{J_s}^T}{\|\boldsymbol{X}_i-\boldsymbol{X}_{i'}\|_2^2}-\mathbf{K}_{J_s,J_s}\Big)\boldsymbol{b}\Big)\\
=&\mathbb{E}\exp\Big(t\cdot\frac{1}{\text{card}(S_n)}\sum_{\sigma\in S_n}\frac{2}{n}\sum_{(i,i')\in O(\sigma)}\boldsymbol{b}^T\Big(\frac{(\boldsymbol{X}_i-\boldsymbol{X}_{i'})_{J_s}(\boldsymbol{X}_i-\boldsymbol{X}_{i'})_{J_s}^T}{\|\boldsymbol{X}_i-\boldsymbol{X}_{i'}\|_2^2}-\mathbf{K}_{J_s,J_s}\Big)\boldsymbol{b}\Big)\\
\le&\frac{1}{\text{card}(S_n)}\sum_{\sigma\in S_n}\mathbb{E}\exp\Big(t\cdot\frac{2}{n}\sum_{(i,i')\in O(\sigma)}\boldsymbol{b}^T\Big(\frac{(\boldsymbol{X}_i-\boldsymbol{X}_{i'})_{J_s}(\boldsymbol{X}_i-\boldsymbol{X}_{i'})_{J_s}^T}{\|\boldsymbol{X}_i-\boldsymbol{X}_{i'}\|_2^2}-\mathbf{K}_{J_s,J_s}\Big)\boldsymbol{b}\Big)\\
=&\mathbb{E}\exp\Big(t\cdot\frac{2}{n}\sum_{(i,i')\in O(\sigma_0)}\boldsymbol{b}^T\Big(\frac{(\boldsymbol{X}_i-\boldsymbol{X}_{i'})_{J_s}(\boldsymbol{X}_i-\boldsymbol{X}_{i'})_{J_s}^T}{\|\boldsymbol{X}_i-\boldsymbol{X}_{i'}\|_2^2}-\mathbf{K}_{J_s,J_s}\Big)\boldsymbol{b}\Big),
\end{aligned} \tag{C.2}$$

where the inequality is due to the Jensen's inequality.

Let $m := n/2$ and recall that $\boldsymbol{X} = (X_1, \ldots, X_d)^T \in \mathcal{M}_d(\boldsymbol{\Sigma}, \xi, q, s; \kappa_L, \kappa_U)$. Letting $\widetilde{\boldsymbol{X}} = (\widetilde{X}_1, \ldots, \widetilde{X}_d)^T$ be an independent copy of $\boldsymbol{X}$, by Equation (B.3), we have that, for any $t \in \mathbb{R}$ and $\boldsymbol{v} \in \mathbb{S}^{d-1}$,

$$\mathbb{E} \exp\left(t \cdot \boldsymbol{v}^T \left(\frac{(\boldsymbol{X} - \widetilde{\boldsymbol{X}})(\boldsymbol{X} - \widetilde{\boldsymbol{X}})}{\|\boldsymbol{X} - \widetilde{\boldsymbol{X}}\|_2^2} - \mathbf{K}\right)\boldsymbol{v}\right) \leq e^{\eta t^2}.$$

In particular, letting $\boldsymbol{v}_{J_s} = \boldsymbol{b}$ and $\boldsymbol{v}_{J_s^C} = \mathbf{0}$, we have

$$\mathbb{E} \exp\left(t \cdot \boldsymbol{b}^T \left(\frac{(\boldsymbol{X} - \widetilde{\boldsymbol{X}})_{J_s}(\boldsymbol{X} - \widetilde{\boldsymbol{X}})_{J_s}^T}{\|\boldsymbol{X} - \widetilde{\boldsymbol{X}}\|_2^2} - \mathbf{K}_{J_s,J_s}\right)\boldsymbol{b}\right) \leq e^{\eta t^2}. \tag{C.3}$$

Then we are able to continue Equation (C.2) as

$$\begin{aligned}
&\mathbb{E} \exp\left(t \cdot \frac{2}{n} \sum_{(i,i') \in O(\sigma_0)} \boldsymbol{b}^T \left(\frac{(\boldsymbol{X}_i - \boldsymbol{X}_{i'})_{J_s}(\boldsymbol{X}_i - \boldsymbol{X}_{i'})_{J_s}^T}{\|\boldsymbol{X}_i - \boldsymbol{X}_{i'}\|_2^2} - \mathbf{K}_{J_s,J_s}\right)\boldsymbol{b}\right) \\
=&\mathbb{E} \exp\left(\frac{t}{m} \sum_{i=1}^m \boldsymbol{b}^T \left(\frac{(\boldsymbol{X}_{2i} - \boldsymbol{X}_{2i-1})_{J_s}(\boldsymbol{X}_{2i} - \boldsymbol{X}_{2i-1})_{J_s}^T}{\|\boldsymbol{X}_{2i} - \boldsymbol{X}_{2i-1}\|_2^2} - \mathbf{K}_{J_s,J_s}\right)\boldsymbol{b}\right) \\
=&\left(\mathbb{E} e^{\frac{t}{m}((\boldsymbol{b}^T S(\boldsymbol{X})_{J_s})^2 - \boldsymbol{b}^T \mathbf{K}_{J_s,J_s}\boldsymbol{b})}\right)^m \leq e^{\eta t^2/m},
\end{aligned} \tag{C.4}$$

where by Equation (B.3), the last inequality holds for any $|t/m| \leq c/\sqrt{\eta}$. Accordingly, choosing $t = \beta m/(2\eta)$, using the Markov inequality, we have

$$\mathbb{P}\left(\boldsymbol{b}^T \left[\widehat{\mathbf{K}} - \mathbf{K}\right]_{J_s,J_s} \boldsymbol{b} > \beta\right) \leq e^{-n\beta^2/(8\eta)}, \quad \text{for } \beta \leq 2c_0\sqrt{\eta}. \tag{C.5}$$

By symmetry, we have the same bound for $\mathbb{P}\left(\boldsymbol{b}^T \left[\widehat{\mathbf{K}} - \mathbf{K}\right]_{J_s,J_s} \boldsymbol{b} < -\beta\right)$ as in Equation (C.5). Together, they give us Equation (C.1). This completes the proof. □